\documentclass[11pt,letterpaper]{article}

\usepackage[margin=1.05in]{geometry}
\usepackage[T1]{fontenc}
\usepackage{lmodern}
\usepackage[round,authoryear]{natbib}
\usepackage{amsmath,amssymb,amsthm,mathtools,bm}
\usepackage{booktabs,multirow,tabularx,array}
\usepackage{graphicx}
\usepackage{placeins}
\usepackage{microtype}
\usepackage[font=small,labelfont=bf]{caption}
\usepackage{enumitem}
\newif\ifScopeStandardAlgorithms
\ScopeStandardAlgorithmstrue
\IfFileExists{algorithm.sty}{}{\ScopeStandardAlgorithmsfalse}
\IfFileExists{algpseudocode.sty}{}{\ScopeStandardAlgorithmsfalse}
\IfFileExists{algorithmicx.sty}{}{\ScopeStandardAlgorithmsfalse}
\usepackage{algorithm}
\usepackage{algpseudocode}
\usepackage[table]{xcolor}
\definecolor{scoperow}{HTML}{E9F1FB}   
\usepackage{url}
\usepackage[hidelinks]{hyperref}
\hypersetup{
  pdftitle={SCOPE Before You Sample: Conformal Scores Guide Posterior Exploration for Fast and Calibrated Bayesian Predictive Inference},
  pdfauthor={Zahra Moslemi, Ziyi Liang, Babak Shahbaba}
}
\setkeys{Gin}{draft=false}
\graphicspath{{figures/}}
\newtheorem{theorem}{Theorem}
\newtheorem{proposition}{Proposition}

\newcommand{\Dtrain}{\mathcal{D}_{\mathrm{train}}}
\newcommand{\Dguide}{\mathcal{D}_{\mathrm{guide}}}
\newcommand{\Dconf}{\mathcal{D}_{\mathrm{conf}}}
\newcommand{\Dtest}{\mathcal{D}_{\mathrm{test}}}
\newcommand{\SCOPE}{SCOPE}
\newcommand{\R}{\mathbb{R}}
\newcommand{\N}{\mathcal{N}}
\newcommand{\E}{\mathbb{E}}
\newcommand{\Prb}{\mathbb{P}}
\newcommand{\SigVI}{\Sigma_{\mathrm{VI}}}
\newcommand{\thetaVI}{\theta_{\mathrm{VI}}}
\newcommand{\thetaMAP}{\theta_{\mathrm{MAP}}}
\newcommand{\thetastar}{\theta_{\star}}
\newcommand{\pSCOPE}{\widehat p_{\mathrm{S}}}
\newcommand{\CSCOPE}{\widehat C_{\alpha}^{\mathrm{S}}}
\newcommand{\lblend}{\lambda_{\mathrm{blend}}}
\newcommand{\TBD}[1]{\textcolor{black!55}{[TBD\ifx\relax#1\relax\else: #1\fi]}}

\newcommand{\numAlphaCovMaxDev}{0.021}

\newcommand{\numAlphaSplitCovMaxDev}{0.016}
\newcommand{\numAnchorAdamStepsHi}{3{,}995}
\newcommand{\numAnchorAdamStepsLo}{3{,}358}
\newcommand{\numAnchorConvTimeHi}{7.7}
\newcommand{\numAnchorConvTimeLo}{5.2}

\newcommand{\numAnchorMapFracAirfoil}{85}

\newcommand{\numAnchorMapFracEnergy}{100}

\newcommand{\numAnchorMapFracYacht}{100}

\newcommand{\numAnchorRmseConvEnergy}{0.54}

\newcommand{\numAnchorRmseConvYacht}{1.18}

\newcommand{\numAnchorRmseShortEnergy}{2.06}

\newcommand{\numAnchorRmseShortYacht}{3.33}

\newcommand{\numAnchorViFracAirfoil}{15}
\newcommand{\numAnchorViFracConcrete}{95}

\newcommand{\numAnchorViFracSupercond}{100}
\newcommand{\numAnchorViFracWine}{95}

\newcommand{\numBaselinePenaltyMax}{0.057}
\newcommand{\numBoundaryEssVsHmcHi}{309}
\newcommand{\numBoundaryEssVsHmcLo}{32}
\newcommand{\numBoundaryEssVsMalaHi}{118}
\newcommand{\numBoundaryEssVsMalaLo}{30}
\newcommand{\numBwCovMaxDev}{0.004}

\newcommand{\numBwNllMaxDev}{0.004}

\newcommand{\numCovAirfoil}{0.891}

\newcommand{\numCovNaval}{0.901}
\newcommand{\numCovProtein}{0.902}

\newcommand{\numEssRatioHmcAnyPercentileMin}{52}
\newcommand{\numEssRatioHmcMedianHi}{647}
\newcommand{\numEssRatioHmcMedianLo}{55}

\newcommand{\numEssRatioHmcMinHi}{1{,}303}
\newcommand{\numEssRatioHmcMinLo}{60}

\newcommand{\numEssRatioHmcPNinetyHi}{412}
\newcommand{\numEssRatioHmcPNinetyLo}{52}

\newcommand{\numFinalPassLoadHi}{1.45}
\newcommand{\numFinalPassLoadLo}{1.08}

\newcommand{\numGuardMedianHi}{1.00}
\newcommand{\numGuardMedianLo}{0.21}

\newcommand{\numGuardNBelowHalf}{6}

\newcommand{\numGuideNllMapSupercond}{6.19}

\newcommand{\numGuideNllViSupercond}{3.85}

\newcommand{\numHOneBoundaryVsPcaHi}{2.9}
\newcommand{\numHOneBoundaryVsPcaLo}{0.9}

\newcommand{\numHOneProbeVsPcaHi}{1.7}
\newcommand{\numHOneProbeVsPcaLo}{0.9}

\newcommand{\numHThreeGapMax}{0.029}
\newcommand{\numHThreeGapMaxAtLeastHundred}{0.023}

\newcommand{\numHThreeGapMin}{-0.005}

\newcommand{\numHThreeNds}{4}
\newcommand{\numHThreeSmallest}{25}
\newcommand{\numHTwoCovDiffMaxAbs}{0.007}

\newcommand{\numHTwoN}{7}
\newcommand{\numHTwoNActiveWorse}{7}

\newcommand{\numHTwoNllGainMax}{0.194}

\newcommand{\numHTwoTvEnergyActive}{0.132}
\newcommand{\numHTwoTvEnergyCompleted}{0.109}

\newcommand{\numHTwoTvYachtActive}{0.213}
\newcommand{\numHTwoTvYachtCompleted}{0.109}
\newcommand{\numHTwoUnguardNllPenaltyMax}{1.35}
\newcommand{\numHTwoUnguardSizeRatioMax}{3.1}
\newcommand{\numHTwoUnguardSizeRatioMaxName}{Yacht}

\newcommand{\numHmcGradients}{945{,}000}
\newcommand{\numHmcLeapfrog}{20}

\newcommand{\numHmcStrongNllShiftMaxAbs}{0.023}

\newcommand{\numHmcTimeMedian}{1{,}171}
\newcommand{\numHmcTrajectories}{45{,}000}

\newcommand{\numKlTrueSimTwoHmc}{0.231}

\newcommand{\numKlTrueSimTwoScope}{0.210}

\newcommand{\numMalaNllGapWorst}{0.771}

\newcommand{\numNAll}{11}

\newcommand{\numNMain}{8}

\newcommand{\numNScopeWithinPointOne}{6}

\newcommand{\numProbeEssVsHmcHi}{87}
\newcommand{\numProbeEssVsHmcLo}{27}
\newcommand{\numProbeEssVsMalaHi}{76}
\newcommand{\numProbeEssVsMalaLo}{25}
\newcommand{\numRankCovMaxDev}{0.010}
\newcommand{\numRankHi}{80.0}
\newcommand{\numRankLo}{3.6}
\newcommand{\numRankNllMaxDev}{0.037}
\newcommand{\numReuseGapMaxAbs}{0.023}

\newcommand{\numReuseSizeRatioHi}{1.054}
\newcommand{\numReuseSizeRatioLo}{0.989}
\newcommand{\numScopeCovHi}{0.918}
\newcommand{\numScopeCovLo}{0.897}

\newcommand{\numScopeMainProposals}{23{,}000}

\newcommand{\numScopeNllGapMedian}{0.002}
\newcommand{\numScopeNllGapWorst}{0.266}

\newcommand{\numScopePilotProposals}{1{,}900}

\newcommand{\numScopeTimeMedianEtoE}{89}

\newcommand{\numSeedCount}{20}

\newcommand{\numSizeBeforeEnergy}{1.89}
\newcommand{\numSizeBeforeNaval}{8.11}

\newcommand{\numSizeBeforeYacht}{2.45}

\newcommand{\numSizeRatioNaval}{1.37}

\newcommand{\numSplitCovHi}{0.911}
\newcommand{\numSplitCovLo}{0.893}

\newcommand{\numSymDiffSimTwoCqr}{3.01}
\newcommand{\numSymDiffSimTwoHmc}{0.811}

\newcommand{\numSymDiffSimTwoScope}{0.817}

\newcommand{\numTvTrueSimTwoHmc}{0.213}

\newcommand{\numTvTrueSimTwoScope}{0.209}

\newif\ifAlphaReady
\AlphaReadytrue
\newif\ifBandwidthReady
\BandwidthReadytrue
\newif\ifCompletionReady
\CompletionReadytrue
\newif\ifFunctionalsReady
\FunctionalsReadytrue
\newif\ifGuideSizeReady
\GuideSizeReadytrue
\newif\ifHmcRerunReady
\HmcRerunReadytrue
\newif\ifHmcStrongReady
\HmcStrongReadytrue
\newif\ifOptionBReady
\OptionBReadytrue
\newif\ifPairedReady
\PairedReadytrue
\newif\ifPowerReady
\PowerReadyfalse
\newif\ifRankReady
\RankReadytrue
\newif\ifSimsSelected
\SimsSelectedtrue
\newif\ifSpanActiveReady
\SpanActiveReadytrue
\newif\ifSpanReady
\SpanReadytrue
\newif\ifSpanSelCurvAllDecoupled
\SpanSelCurvAllDecoupledfalse
\newif\ifSpanSelOtherMoreBnd
\SpanSelOtherMoreBndtrue
\newif\ifSpanSelOtherMoreBndPay
\SpanSelOtherMoreBndPaytrue
\newif\ifSpanSelReady
\SpanSelReadytrue
\newif\ifSpanSelWeightLose
\SpanSelWeightLosetrue
\newif\ifSpanSelWeightLoseIsVi
\SpanSelWeightLoseIsVitrue
\newif\ifSplitReady
\SplitReadytrue

\newif\ifSimTwoSpanReady
\SimTwoSpanReadytrue
\newcommand{\numRecordedAnchorRankSelected}{4.8}
\newcommand{\numRecordedAnchorRankEnvelope}{4.7}

\providecommand{\numPessScopeMedian}{30}
\providecommand{\numPessChainMedian}{20}
\providecommand{\numPessWorstMedian}{106}
\providecommand{\numPessChainWorstMedian}{72}
\providecommand{\numPessMalaMedian}{0.85}
\providecommand{\numPessMalaWorstMedian}{1.3}

\newcommand{\numSimGuidanceRank}{4}
\newcommand{\numSimOneGuidedErr}{13.0}
\newcommand{\numSimOneUniformErr}{16.7}
\newcommand{\numSimOneDiff}{-3.70}
\newcommand{\numSimOneDiffLo}{-5.6}
\newcommand{\numSimOneDiffHi}{-1.8}

\newcommand{\numSimTwoGuidedErrRfour}{37.0}
\newcommand{\numSimTwoUniformErrRfour}{47.4}
\newcommand{\numSimTwoDiff}{-10.4}
\newcommand{\numSimTwoDiffLo}{-15.6}
\newcommand{\numSimTwoDiffHi}{-5.1}

\newcommand{\numGuidanceSweepRTwo}{30}
\newcommand{\numGuidanceSweepRFour}{22}
\newcommand{\numGuidanceSweepREight}{16}
\newcommand{\numGuidanceSweepRSixteen}{9}
\newcommand{\numGuidanceSweepRThirtyTwo}{3}

\title{SCOPE Before You Sample: Conformal Scores Guide Posterior Exploration for Fast and Calibrated Bayesian Predictive Inference}
\author{%
Zahra Moslemi$^{1}$ \qquad Ziyi Liang$^{2}$ \qquad Babak Shahbaba$^{1}$\\[0.8ex]
\normalsize $^{1}$University of California, Irvine\\
\normalsize $^{2}$University of California, Santa Barbara
}
\date{}

\begin{document}
\raggedbottom
\maketitle
\begin{abstract}
Bayesian predictive regions account for parameter uncertainty but can be
poorly calibrated under model misspecification, even with expensive full
MCMC. Cheaper approximations such as variational inference reduce the cost but often yield less accurate predictive distributions. 
We propose SCOPE (Score-guided Conformal Posterior Exploration), which retains the predictive quality of full MCMC while achieving calibrated coverage at a fraction of its cost. The key observation is that a prediction region can depend on the posterior mainly through a few parameter directions that move the region's boundary, and these are generally not the directions of largest posterior variance. 
SCOPE gives nonconformity scores a new role in guiding posterior computation: score gradients on a held-out guide set, weighted toward the target boundary, identify the relevant directions. MCMC explores this subspace, while an inexpensive projected variational approximation carries the remaining uncertainty. The same guide set calibrates the region’s final density level. We establish asymptotic coverage for SCOPE and provide a split variant with finite-sample validity. Controlled simulations show that score-guided exploration yields more accurate calibrated regions than non-targeted direction choices at low active rank. Across simulated and real benchmarks, SCOPE has a median test-NLL gap of only 0.002 nats relative to full Hamiltonian Monte Carlo (HMC), and on the real benchmarks it runs 12--15$\times$ faster.
\end{abstract}

\section{Introduction}
\label{sec:introduction}

Bayesian posterior predictive distributions account for parameter uncertainty,
but computing them with full-dimensional Markov chain Monte Carlo (MCMC),
including Hamiltonian Monte Carlo (HMC) \citep{neal2011hmc}, can be
computationally demanding. Variational inference
\citep{blei2017variational} and Laplace approximations
\citep{daxberger2021laplace} reduce this cost, but may alter the predictive
distribution. Moreover, exact posterior computation does not itself guarantee
frequentist coverage: under model misspecification, Bayesian prediction
regions can under- or over-cover
\citep{syring2019calibrating,fong2021conformalbayes}.
With a fitted predictor and a genuinely independent calibration sample,
split-conformal methods give finite-sample marginal coverage under
exchangeability \citep{vovk2005algorithmic,lei2018distributionfree}.
Conformal calibration adjusts a region's threshold, however; it cannot by
itself repair an inaccurate predictive density or recover missing features
such as multimodality.

The downstream target suggests a more economical way to spend Bayesian
computation. A prediction region can be strongly affected by a small set of
parameter directions that move its boundary, even if those directions are
not the dominant directions of posterior variance. Rather than improving a
global posterior approximation indiscriminately, we seek to use expensive
sampling where its effect on the reported region is largest.

In density-based split-conformal prediction \citep{izbicki2022hpd},
given a fitted predictive density $\widehat p(y\mid x)$, one uses
$-\log \widehat p(y\mid x)$ as the nonconformity score, and a
response is included when its score falls below a calibrated threshold. We give this score a new role: guiding posterior computation. The intuition
is that changing a score near the threshold can move a response into or
out of the region, whereas the same change far from the threshold may
leave membership unchanged. We therefore emphasize score gradients on
held-out observations near the threshold to identify parameter directions
that can affect region membership, accounting for local posterior curvature.

\begin{figure}[t]
    \centering
    \includegraphics[width=\textwidth]{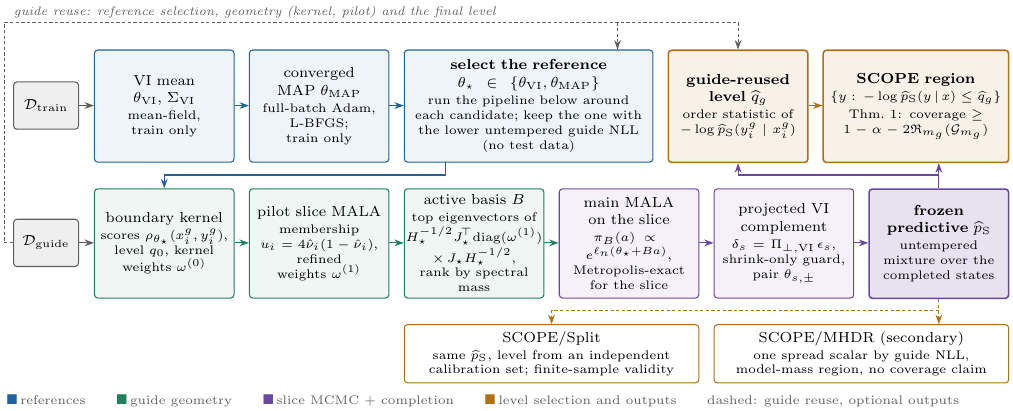}
    \caption{\textbf{SCOPE overview.} For each candidate reference point,
    guide observations determine an active parameter subspace.
    Slice MCMC and projected variational completion produce a candidate
    predictive distribution. The candidate with the lower negative
    log-likelihood (NLL) on the guide set is selected, and a quantile of
    the guide scores determines its prediction region.
    \SCOPE/Split uses an independent calibration set for the final
    threshold.}
    \label{fig:scope-overview}
\end{figure}

\SCOPE\ (\emph{Score-guided Conformal Posterior Exploration}) runs MCMC
in this subspace, while an inexpensive projected variational approximation
carries the remaining uncertainty.
The same held-out set then calibrates the final regions. Figure~\ref{fig:scope-overview} overviews the proposed pipeline.
There is an important distinction between the two calibration options.
Default \SCOPE\ reuses guide observations when selecting its subspace and
calibrating the final threshold. Because the scores then depend on those
observations, ordinary split-conformal finite-sample validity does not follow.
Our analysis instead bounds the guide-reuse effect through region-class
complexity. When finite-sample marginal validity is essential, \SCOPE/Split
uses a separate calibration set that is untouched by subspace selection and
anchor choice.

Our main contributions are as follows.
\begin{enumerate}[leftmargin=1.35em,itemsep=0.3mm,topsep=0.4mm]
    \item \textbf{Region-directed allocation of posterior computation.}
    We use boundary-localised nonconformity-score gradients to select the
    active MCMC subspace. At fixed rank, this span maximises local
    boundary-weighted score sensitivity; controlled simulations show lower
    oracle-region error than uniform score weighting, with larger reductions at smaller fixed ranks.

    \item \textbf{A practical sampler.}
    \SCOPE\ confines MCMC to a region-sensitive, low-dimensional subspace,
    allowing a smaller chain budget in our evaluated settings. A projected
    variational approximation represents the remaining uncertainty without
    requiring another Markov chain.    

    \item \textbf{Coverage guarantees.}
   For \SCOPE, which reuses the same held-out data for guidance and calibration,
    we establish a coverage bound yielding asymptotic marginal validity. We also propose \SCOPE/Split, which provides
    finite-sample marginal validity with independent calibration.

    \item \textbf{Empirical evidence.}
    Across two simulations and six main UCI datasets, \SCOPE\ has a median
    signed test-NLL difference of $\numScopeNllGapMedian$ nats relative to
    the specified fixed-budget, full-dimensional HMC baseline, with a
    worst-case excess of
    $\numScopeNllGapWorst$ nats. On the six UCI datasets, median runtime is
    $\numScopeTimeMedianEtoE$ seconds for \SCOPE\ versus
    $\numHmcTimeMedian$ seconds for HMC.
\end{enumerate}

\section{Related Work}
\label{sec:related}

\paragraph{Reduced-space Bayesian inference.}
Reduced-space methods concentrate posterior computation in a few
parameter directions. Likelihood-informed subspaces use posterior-averaged,
prior-preconditioned Gauss--Newton curvature and approximate the
posterior using a projected likelihood with a Gaussian prior
complement \citep{cui2014likelihood}.
Active subspaces use gradient information to identify influential
directions of variation in a scalar function \citep{constantine2015active}.
Active-subspace MCMC uses prior-averaged outer products of misfit
gradients, approximates the misfit by its conditional prior average,
and samples inactive variables from the conditional prior
\citep{constantine2016accelerating}.
For neural networks, subspace inference places priors on coordinates
within affine spans, including PCA spans of SGD trajectories
\citep{izmailov2020subspace}, while subnetwork inference restricts
uncertainty to selected parameters \citep{daxberger2021subnetwork}.
SWAG instead fits a low-rank-plus-diagonal Gaussian to an SGD trajectory
\citep{maddox2019swag}.

More broadly, loss-calibrated inference and goal-oriented dimension
reduction adapt inference to downstream decisions or quantities of
interest
\citep{lacostejulien2011loss,kusmierczyk2019variational,spantini2017goal}.

\SCOPE\ shares the task-aware perspective but targets a prediction region
at a specified coverage level. Rather than optimising a decision loss or
preserving a predefined quantity of interest, it selects directions
according to their local effect on scores near the region boundary. Consequently, changing the target coverage
level can change which parameter directions receive MCMC computation,
even when the underlying posterior remains fixed. Matched span controls
assess this choice (Sections~\ref{sec:where} and~\ref{sec:h1}).

\paragraph{Conformal prediction and predictive densities.}
Conformal prediction \citep{vovk2005algorithmic} turns any predictive model
into prediction sets with marginal coverage.
CQR adapts intervals to heteroscedasticity
\citep{romano2019cqr}; CD-split and HPD-split construct density-based,
potentially disconnected regions \citep{izbicki2022hpd}. Conformal Bayesian
Computation applies full-conformal reasoning to Bayesian predictive densities
\citep{fong2021conformalbayes}.
Conformal criteria have also been used to influence predictive-model training
\citep{stutz2022learning,einbinder2022training}.
The distinctive step in \SCOPE\ is to use boundary-weighted gradients of
nonconformity scores to choose the subspace in which posterior MCMC is run,
rather than using conformal scores only after fitting or as a training loss.

\paragraph{Calibration with data-dependent scores.}
The standard split-conformal rank guarantee presumes that the calibration
data do not influence the fitted score function. Alternative constructions,
such as jackknife+, use data reuse with different exchangeability arguments
\citep{barber2021jackknife}; selecting a model or its tuning parameters using
calibration data can also change conformal validity
\citep{liang2024selection,zeng2025tuning}. Default \SCOPE\ uses guide labels
for geometry, anchor selection, and the final cutoff, so it needs a
reuse-aware bound. \SCOPE/Split instead calibrates on fresh data, preserving
the standard finite-sample marginal guarantee.

\section{Problem Setup}
\label{sec:background}

Let $p_\theta(y\mid x)$, $\theta\in\R^D$, be a conditional
density model with prior $p_0(\theta)$. The training set
$\Dtrain=\{(x_i,y_i)\}_{i=1}^{n}$ defines the posterior
\begin{equation}
\begin{aligned}
\pi_n(\theta)&\propto \exp\{\ell_n(\theta)\},\\
\ell_n(\theta)
&=\log p_0(\theta)
+\textstyle\sum_{i=1}^{n}\log p_\theta(y_i\mid x_i),
\end{aligned}
\label{eq:posterior}
\end{equation}
and the posterior predictive density
\[
p_n(y\mid x)
=\int p_\theta(y\mid x)\pi_n(\theta)\,d\theta.
\]

\paragraph{Data roles.}
Only $\Dtrain$ enters the posterior~\eqref{eq:posterior}.
A disjoint guide set
$\Dguide=\{(x_i^g,y_i^g)\}_{i=1}^{m_g}$
tells \SCOPE\ where to concentrate posterior computation and, in the
default guide-reused construction, also sets the final region threshold.
It influences the approximation but not the posterior target.
A test set $\Dtest$ is used only for evaluation.

\begin{figure}[!tb]
\centering
\includegraphics[width=0.9\columnwidth]{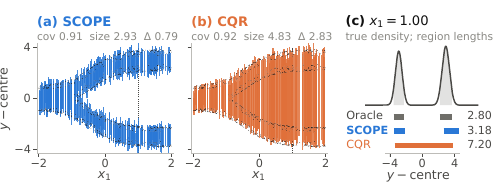}
\caption{\textbf{Density-level geometry in Simulation~2} (one seed).
(a,b)~\SCOPE\ and CQR regions versus $x_1$; dots mark population global-level
oracle boundaries. (c)~At a disconnected input, \SCOPE\ follows both modes
whereas CQR spans the low-density gap.}
\label{fig:sim2}
\end{figure}

\paragraph{Nonconformity scores and density-level regions.}
We use two related nonconformity scores. For a single parameter value,
\[
\rho_\theta(x,y)=-\log p_\theta(y\mid x)
\]
measures how unusual a response is under the model with parameters
$\theta$; its gradient with respect to $\theta$ is what guides posterior
exploration in Section~\ref{sec:method}. For a fitted predictive density
$\widehat p$, the corresponding score is
\[
R_{\widehat p}(x,y)=-\log\widehat p(y\mid x),
\]
and it is this score that calibrates the final region. Thresholding $R_{\widehat p}$ gives
the density-level region
\[
C_q(x)
=\{y:R_{\widehat p}(x,y)\le q\}
=\{y:\widehat p(y\mid x)\ge e^{-q}\},
\]
so calibration amounts to choosing the density level $e^{-q}$. Unlike
intervals, such regions can be disconnected and can adapt to
heteroscedastic or multimodal predictive distributions. Figure~\ref{fig:sim2} illustrates why this geometry matters.
In Simulation~2, the population global-level oracle region changes between
connected and disconnected sets as the input varies. \SCOPE's density-level
region can follow this change, whereas an interval-valued method such as CQR
must span the low-density gap between modes.

\section{SCOPE}
\label{sec:method}

\SCOPE\ constructs a candidate predictive around each of two
training-fitted anchors (Figure~\ref{fig:scope-overview}).
Guide-score gradients select the active subspace
(Section~\ref{sec:where}); slice MCMC and variational completion
form each predictive (Section~\ref{sec:how}). Guide NLL then
selects the predictive to retain (Section~\ref{sec:anchor}), after
which a quantile of its guide scores sets the region threshold
(Section~\ref{sec:level}). Section~\ref{sec:validity} gives the
coverage results, and Algorithm~\ref{alg:scope} summarises the
procedure.

\paragraph{Candidate anchors.}
We construct two candidate anchors from $\Dtrain$. A diagonal Gaussian
variational approximation $q_\phi(\theta)=\N(\mu_\phi,\SigVI)$,
$\SigVI=\operatorname{diag}(\sigma_{\phi,1}^2,\ldots,\sigma_{\phi,D}^2)$, is
fit by maximising the evidence lower bound
\citep{blei2017variational}, giving $\thetaVI=\mu_{\widehat\phi}$. The second
candidate, $\thetaMAP$, continues MAP optimisation to a locally converged
posterior mode (Appendix~\ref{app:anchor}). Whichever anchor is selected,
$\SigVI$ supplies the scales of the residual uncertainty in
Section~\ref{sec:how}. Until Section~\ref{sec:anchor}, $\thetastar$ denotes
either candidate.

\subsection{Where to Sample: Score-Guided Active Subspace}
\label{sec:where}

Using the per-parameter score $\rho_\theta$ of
Section~\ref{sec:background}, at the anchor let
$s_i^{(0)}=\rho_{\thetastar}(x_i^g,y_i^g)$ and
$q_0=\operatorname{Quantile}_{1-\alpha}\{s_i^{(0)}\}$. A smooth kernel
emphasizes guide scores near this level,
\begin{equation}
\omega_i^{(0)}
\propto
\exp\!\left\{
-\frac{(s_i^{(0)}-q_0)^2}{2h^2}
\right\},
\label{eq:kernel}
\end{equation}
where the bandwidth is $h=\max\{c_h\operatorname{MAD}(s^{(0)}),h_{\min}\}$,
$\operatorname{MAD}(s^{(0)})=\operatorname{median}_i|s_i^{(0)}-\operatorname{median}_js_j^{(0)}|$
is the median absolute deviation of the guide scores, $c_h>0$ is a scale
constant and $h_{\min}>0$ is a floor for nearly constant scores; the weights
are normalized to sum to one.
Guide points with scores near $q_0$ lie close to the
boundary of the region, so small parameter changes can move them in or out of
it; points far from $q_0$ are firmly inside or outside and say little about
where the boundary lies. Let
$J_\star\in\R^{m_g\times D}$ stack the guide-score gradients
$\nabla_\theta\rho_{\thetastar}(x_i^g,y_i^g)^\top$, and let $H_\star\succ0$ be
a positive-definite curvature metric at $\thetastar$ (defined in
Appendix~\ref{app:hstar}). Whitening by $H_\star^{-1/2}$ measures each
direction in units of local posterior spread, so a direction counts as
important only if the posterior allows appreciable movement along it
\emph{and} that movement changes the scores. For normalized nonnegative
weights $\omega$ define
\begin{equation}
M_\star(\omega)
=
H_\star^{-1/2}
J_\star^\top\operatorname{diag}(\omega)J_\star
H_\star^{-1/2}.
\label{eq:operator}
\end{equation}

\begin{proposition}[Local weighted score sensitivity]
\label{prop:local-relevance}
Suppose the guide scores are differentiable at $\thetastar$.
Let $v\in\R^D$, $\|v\|_2=1$, and
$\delta\theta=H_\star^{-1/2}v$. As $t\to0$,
\begin{align}
&\sum_{i=1}^{m_g}
\omega_i
\left[
\rho_{\thetastar+t\delta\theta}(x_i^g,y_i^g)
-
\rho_{\thetastar}(x_i^g,y_i^g)
\right]^2
\nonumber\\
&\hspace{18mm}
=
t^2v^\top M_\star(\omega)v+o(t^2).
\label{eq:firstorder}
\end{align}
Hence any direction with $J_\star H_\star^{-1/2}v=0$ has zero first-order
weighted guide-score motion, irrespective of posterior variance in that
direction. Moreover, for every $r$, the $r$ leading eigenvectors of
$M_\star(\omega)$ span an $r$-dimensional subspace of whitened perturbations
with the largest total local weighted score sensitivity, equal to the sum of
the $r$ largest eigenvalues.
\end{proposition}

In words, score guidance is a resource-allocation rule. For a fixed active
dimension it assigns the non-Gaussian sampling budget to the subspace that
captures the greatest local boundary-weighted score variation. The criterion
is local and first order (Appendix~\ref{app:prop1}). With the notation above,
the active basis is
\begin{equation}
B_r=H_\star^{-1/2}P_r,
\qquad
B_r^\top H_\star B_r=I_r,
\label{eq:basis}
\end{equation}
where $r$ is the smallest rank satisfying
$\sum_{j=1}^{r}\lambda_j\ge\eta\sum_j\lambda_j$ for the positive spectrum,
subject to fixed lower and upper bounds on $r$. By
Proposition~\ref{prop:local-relevance}, before these bounds are applied no
curvature-orthonormal subspace of smaller dimension can capture this fraction
$\eta$ of the local weighted score sensitivity. Thus the spectral-mass rule
uses the fewest sampled directions that attain its prescribed local capture
level. The values of $c_h$, $h_{\min}$, $\eta$, the rank bounds and
$\lblend$ in~\eqref{eq:blend} are fixed across all experiments and listed in
Appendix~\ref{app:hyper}.

\paragraph{Membership refinement.}
The kernel weights $\omega^{(0)}$ judge closeness to the boundary at the
anchor alone. What matters, however, is which guide points actually move in
or out of the region as the parameters vary under the posterior. We
therefore check this directly with a short pilot chain. We first build a
provisional basis $B_0$ from $\omega^{(0)}$ and run a short MALA chain
targeting $\pi_0(a)\propto\exp\{\ell_n(\thetastar+B_0a)\}$. For each
retained state $\theta_t^{\mathrm{pil}}=\thetastar+B_0a_t$, we recompute the
boundary $q_t$ and record whether each guide point lies inside the region.
The fraction of states in which point $i$ is inside,
$\widehat\nu_i=N_0^{-1}\sum_t\mathbf 1\{\rho_{\theta_t^{\mathrm{pil}}}(x_i^g,y_i^g)\le q_t\}$,
gives the membership variance $u_i=4\widehat\nu_i(1-\widehat\nu_i)\in[0,1]$. We blend it with the anchor-level
weights,
\begin{equation}
\omega_i^{(1)}
=
(1-\lblend)
\frac{u_i}{\sum_ju_j}
+
\lblend\,\omega_i^{(0)},
\label{eq:blend}
\end{equation}
and revert to $\omega^{(0)}$ if all $u_i$ are numerically zero. The final
basis $B$ is then rebuilt from $M_\star(\omega^{(1)})$ and frozen. The pilot
changes only the weights; the gradients in $J_\star$ are still evaluated at
$\thetastar$. Because these are full-parameter gradients, the refined span
need not lie within the provisional one.
\paragraph{From local guidance to the final region.}
Proposition~\ref{prop:local-relevance} characterises a local allocation
objective, but it does not by itself show that the resulting directions improve
the final prediction region. We therefore test this downstream consequence
directly on both main simulations. At a fixed active rank, we compare completed
SCOPE constructions using boundary-sensitive and uniform score weighting.
Within each comparison, the training-fitted anchor, score Jacobian, curvature
metric, main-chain proposal budget and adaptation, projected VI completion,
energy guard, and independent calibration sample are held fixed; only the
weighting of guide observations used to select the active span changes. This
design isolates whether concentrating the sampled directions on observations
near the target boundary improves the final independently calibrated region,
rather than merely increasing the local sensitivity criterion of
Proposition~\ref{prop:local-relevance}.

\begin{table}[t]
\centering
\caption{\textbf{Matched rank-$\numSimGuidanceRank$ score-guidance comparison.}
Normalized symmetric-difference error (\% of oracle-region length) to the
population global-level oracle. Guided and uniform differ only in score
weighting; VI/MAP anchor errors are averaged within seed. Difference is
guided$-$uniform with multiplicity-adjusted paired 95\% bootstrap CI over
$\numSeedCount$ seeds.}
\label{tab:bwc-main}
\tiny
\setlength{\tabcolsep}{12.0pt}
\resizebox{\columnwidth}{!}{%
\begin{tabular}{@{}lrrl@{}}
\toprule
& Guided & Uniform & Guided $-$ uniform [95\% CI] \\
\midrule
Simulation~1 & \textbf{\numSimOneGuidedErr} & \numSimOneUniformErr &
$\numSimOneDiff\;[\numSimOneDiffLo,\numSimOneDiffHi]$ \\
Simulation~2 & \textbf{\numSimTwoGuidedErrRfour} & \numSimTwoUniformErrRfour &
$\numSimTwoDiff\;[\numSimTwoDiffLo,\numSimTwoDiffHi]$ \\
\bottomrule
\end{tabular}}
\end{table}

\subsection{How to Sample: Slice MCMC and Variational Completion}
\label{sec:how}

The main chain targets the posterior restricted to the selected affine slice,
\begin{equation}
\begin{aligned}
\pi_B(a)&\propto e^{\ell_n(\thetastar+Ba)},\\
\nabla_a\log\pi_B(a)&=B^\top\nabla_\theta\ell_n(\thetastar+Ba),
\end{aligned}
\label{eq:slice}
\end{equation}
using the Metropolis-adjusted Langevin algorithm
\citep[MALA;][]{roberts1996langevin}. The Metropolis correction preserves this
\emph{slice posterior}; it does not target the full posterior or the
marginal posterior of the active coordinates. Let $a_1,\ldots,a_{N_s}$ be the
retained states and $\theta_s^{\mathrm{act}}=\thetastar+Ba_s$.

On its own, the chain leaves every direction outside the slice fixed at the
anchor and therefore understates the uncertainty. A draw from the VI Gaussian
would supply this missing uncertainty, but it would also perturb the active
directions, which the chain has already sampled, and so count their uncertainty twice. We therefore project the VI draw onto the remaining directions. 
Define $C_B=B^\top\SigVI^{-1}B$, the VI precision restricted to the
active coordinates, and
$\Pi_{\perp,\mathrm{VI}}=I_D-BC_B^{-1}B^\top\SigVI^{-1}$, the projector
that removes the active component in the $\SigVI^{-1}$ metric.
Draw a full-space VI perturbation $\epsilon_s\sim\N(0,\SigVI)$ and
retain its residual component
$\delta_s=\Pi_{\perp,\mathrm{VI}}\epsilon_s$. Then
$B^\top\SigVI^{-1}\delta_s=0$, so the residual is orthogonal to the
active span in this metric, and
$\operatorname{Cov}(\delta_s)=\SigVI-BC_B^{-1}B^\top$ is the VI
covariance with the active contribution removed.


\begin{proposition}[Anchor-centred Gaussian consistency]
\label{prop:gaussian}
Let $a_G\in\R^r$ denote Gaussian active coordinates with
$a_G\sim\N(0,C_B^{-1})$, corresponding to the active component of an
anchor-centred Gaussian draw with covariance $\SigVI$. Independently, let
$\delta\sim\N(0,\SigVI-BC_B^{-1}B^\top)$. Then
$\thetastar+Ba_G+\delta\sim\N(\thetastar,\SigVI)$: the fitted VI Gaussian when
$\thetastar=\thetaVI$, and $\N(\thetaMAP,\SigVI)$, the VI covariance centred at
the MAP, when $\thetastar=\thetaMAP$.
\end{proposition}

Proposition~\ref{prop:gaussian} verifies covariance recovery for the stated
Gaussian active law and unguarded complement. \SCOPE\ instead uses slice-MCMC
active states and energy-based residual shrinkage:
$\widetilde\delta_s=\gamma_s\delta_s$, $0\le\gamma_s\le1$
(Appendix~\ref{app:guard}). The guard changes the residual distribution and
can substantially shrink it; the reported guard-scale summary is $0.21$ on
Energy and Yacht (Table~\ref{tab:app-diag}). The resulting approximation is
assessed by the completion ablation (Appendix~\ref{app:completion}).

Each active state is combined with the complement draw in both directions,
$\theta_{s,\pm}=\theta_s^{\mathrm{act}}\pm\widetilde\delta_s$. These
antithetic pairs keep the complement centred and reduce the Monte Carlo
error of the average below. The resulting predictive density is
\begin{equation}
\begin{aligned}
&\pSCOPE(y\mid x;\Dguide)\\
&\quad=
\frac1{2N_s}
\sum_{s=1}^{N_s}
\left[
p_{\theta_{s,+}}(y\mid x)
+
p_{\theta_{s,-}}(y\mid x)
\right].
\end{aligned}
\label{eq:predictive}
\end{equation}
The pairs are used only to evaluate this average; they are not additional
states of the Markov chain.

\paragraph{Computational cost.}
Each active proposal evaluates a full posterior gradient. The observed
inference-time savings come from the smaller proposal budget used by the
slice chains and completion without an additional chain. They depend on
the accuracy achieved under the specified sampler settings and budgets
(Appendix~\ref{app:computation}).

\subsection{Selecting the Anchor}
\label{sec:anchor}

For $\kappa\in\{\mathrm{VI},\mathrm{MAP}\}$, let
$\widehat p_\kappa$ denote the predictive density
\eqref{eq:predictive} obtained by running
Sections~\ref{sec:where}--\ref{sec:how} around
$\theta_\kappa$. We choose
\begin{equation}
\kappa^\star
=
\arg\min_{\kappa\in\{\mathrm{VI},\mathrm{MAP}\}}
\left[
-\frac{1}{m_g}
\sum_{i=1}^{m_g}
\log \widehat p_\kappa(y_i^g\mid x_i^g)
\right],
\label{eq:anchor-selection}
\end{equation}
and set
$\thetastar=\theta_{\kappa^\star}$ and
$\pSCOPE=\widehat p_{\kappa^\star}$.
Thus we retain the candidate predictive with the lower guide NLL.
The log score is a proper scoring rule
\citep{gneiting2007strictly}; selection uses no test data, and both
candidate fits enter \SCOPE's end-to-end runtime.

\subsection{Setting the Region Threshold}
\label{sec:level}

After anchor selection, freeze the predictive density
$\pSCOPE=\widehat p_{\kappa^\star}$. Each guide observation
contributes one unweighted score
\[
R_i=-\log \pSCOPE(y_i^g\mid x_i^g).
\] 
With
$R_{(1)}\le\cdots\le R_{(m_g)}$, $R_{(m_g+1)}=+\infty$ and
$k_g=\lceil(m_g+1)(1-\alpha)\rceil$, set $\widehat q_g=R_{(k_g)}$. The default
\SCOPE\ region is
\begin{equation}
\CSCOPE(x)
=
\left\{
y:
-\log
\pSCOPE(y\mid x;\Dguide)
\le
\widehat q_g
\right\}.
\label{eq:main-region}
\end{equation}
The geometry weights do not enter this order statistic. Reusing the guide set
avoids a further data split but requires the validity conditions stated in Theorem \ref{thm:reuse}.

\subsection{Guide Reuse and Validity}
\label{sec:validity}

The guide labels determine the geometry and pilot refinement, select the
anchor, and set the prediction threshold. Because the same observations
influence both the fitted score and its calibration, the usual
split-conformal rank argument does not apply to the default
\emph{guide-reused region}. This is related to, but distinct from, data
reuse in jackknife+ \citep{barber2021jackknife} and the effects of model
selection or hyperparameter tuning on conformal coverage
\citep{liang2024selection,zeng2025tuning}. We instead control coverage loss
through the complexity of the family of possible prediction regions.

\begin{theorem}[Reuse-aware marginal coverage]
\label{thm:reuse}
All probabilities and expectations are conditional on $\Dtrain$ and data-independent algorithmic randomness. Fix $\alpha\in(0,1)$. Suppose the observations in $\Dguide$ and a future observation $(X_{\mathrm{new}},Y_{\mathrm{new}})$ are i.i.d.
Let $\mathcal R_{m_g}$ be a class of measurable real-valued score functions
fixed independently of $\Dguide$. Assume that the fitted score
$(x,y)\mapsto-\log\pSCOPE(y\mid x;\Dguide)$ belongs to
$\mathcal R_{m_g}$ almost surely.
Define
\[
\mathcal G_{m_g}=
\left\{
(x,y)\mapsto\mathbf 1\{R(x,y)\le t\}:
\begin{array}{l}
R\in\mathcal R_{m_g},\\
t\in\R\cup\{+\infty\}
\end{array}
\right\}.
\]
Define the expected
Rademacher complexity
\[
\mathfrak R_{m_g}(\mathcal G_{m_g})
=
\E_{\Dguide,\sigma}
\left[
\sup_{g\in\mathcal G_{m_g}}
\frac1{m_g}\sum_{i=1}^{m_g}\sigma_i g(x_i^g,y_i^g)
\right],
\]
where $\sigma_i \overset{\mathrm{i.i.d.}}{\sim} \operatorname{Unif}\{-1,1\}$. Then \eqref{eq:main-region} satisfies
\begin{equation}
\Prb\{Y_{\mathrm{new}}\in\CSCOPE(X_{\mathrm{new}})\}
\ge 1-\alpha-2\mathfrak R_{m_g}(\mathcal G_{m_g}).
\label{eq:reuse-bound}
\end{equation}
\end{theorem}

\paragraph{Class complexity for \SCOPE.}
Class complexity depends on both \SCOPE's guide-dependent choices and the
density family $p_\theta(y\mid x)$. For neural density models with Gaussian
or finite Gaussian-mixture likelihoods and standard activations,
Appendix~\ref{sec:scope-complexity} constructs a guide-independent containing
class and establishes polynomial growth of its membership patterns under
the stated assumptions. With architecture, allowed ranks and $N_s$ fixed, choosing this
containing class as $\mathcal G_{m_g}$ gives
\[
\mathfrak R_{m_g}(\mathcal G_{m_g})
=O\!\left(\sqrt{\frac{\log m_g}{m_g}}\right)\to0.
\]
The coverage penalty therefore vanishes, yielding asymptotic marginal
coverage of at least $1-\alpha$.

\paragraph{\SCOPE/Split: independent calibration.}
For a finite-sample marginal guarantee, reserve an additional calibration set
\[
\Dconf=\{(x_i^c,y_i^c)\}_{i=1}^{m_c}
\]
that is not used to construct the predictive density. After fitting \SCOPE\
using $\Dtrain$ and $\Dguide$, freeze $\widehat p_S$ and compute the
calibration scores
\[
R_i^c=-\log \widehat p_S(y_i^c\mid x_i^c),
\qquad i=1,\ldots,m_c.
\]
Let $R_{(j)}^c$ denote the $j$th smallest calibration score and set
\[
\widehat q_{\rm conf}
=
R^c_{(\lceil(m_c+1)(1-\alpha)\rceil)},
\qquad
R^c_{(m_c+1)}=+\infty.
\]
The resulting split-calibrated region is
\[
\widehat C^{\rm split}_\alpha(x)
=
\{y:-\log \widehat p_S(y\mid x)\le \widehat q_{\rm conf}\}.
\]

\begin{proposition}[Finite-sample validity under independent calibration]
\label{prop:split-validity}
Suppose $\widehat p_S$ is fixed independently of $\Dconf$ and that the
$m_c$ calibration observations and a future observation are exchangeable
conditional on everything used to construct $\widehat p_S$. Then
\[
\Pr\!\left\{
Y_{\mathrm{new}}
\in
\widehat C^{\mathrm{split}}_\alpha(X_{\mathrm{new}})
\right\}
\ge 1-\alpha.
\]
\end{proposition}

This is the usual split-conformal rank guarantee
\citep{lei2018distributionfree}. Thus Theorem~\ref{thm:reuse}
governs the default guide-reused region, whereas \SCOPE/Split has ordinary
finite-sample marginal validity under exchangeability.
\section{Results}
\label{sec:results}

We evaluate \SCOPE\ on two simulations and six main UCI regression
benchmarks using $\numSeedCount$ matched seeds. The simulations have known
data-generating laws, allowing comparison with population global-level oracle
regions. We study score-guided direction selection (H1), variational
completion (H2), and guide reuse (H3), and compare predictive accuracy,
sampling efficiency, and calibrated prediction regions with Bayesian and
conformal baselines. Within each benchmark, Bayesian methods share the model,
data splits, and training-only initialisation; UCI data use
$50/20/15/15\%$ train/guide/calibration/test splits, and regions target
$90\%$ marginal coverage. Finite-sample set comparisons use independent
calibration, while \SCOPE\ otherwise denotes the guide-reused region
\eqref{eq:main-region} of the predictive density
\eqref{eq:predictive}. HMC denotes the specified fixed-budget, full-dimensional
baseline of Appendix~\ref{app:baselines}; timings compare these implementations
and budgets. Score guidance is tested against simulation oracle regions, while
the HMC comparisons assess predictive scores and inference cost. Headline
runtime and ESS/s summaries use six UCI datasets, charging both SCOPE candidates
and selection, and one selected baseline fit. Shared initialisation and test
evaluation are excluded. Full settings and per-benchmark results are in
Appendices~\ref{app:details} and~\ref{app:full-tables}.

\begin{figure}[!t]
\centering
\includegraphics[width=\textwidth]{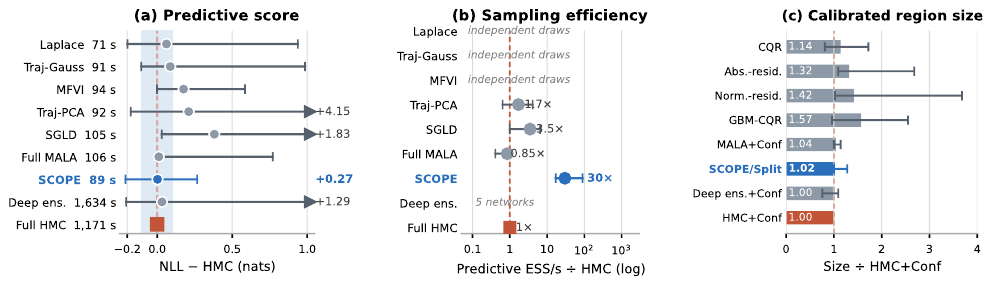}
\vspace{-7pt}
\caption{\textbf{Main empirical summary.}
(a)~Test-NLL excess relative to the HMC baseline
across eight benchmarks (labels: median UCI runtime).
(b)~Predictive ESS/s relative to the HMC baseline on six UCI datasets.
(c)~Independently calibrated region size relative to HMC+Conf across eight
benchmarks; \SCOPE/Split is shown so panel~(c) has standard
split-conformal marginal guarantee. Points/bars are medians and whiskers are
ranges. \vspace{-12pt}}
\label{fig:headline}
\end{figure}

\subsection{Score Guidance Matters Most at Low Active Rank (H1)}
\label{sec:h1}

Score guidance is intended to allocate a limited active sampling budget to the
directions that matter for the reported region. Proposition~\ref{prop:local-relevance}
shows that, at fixed rank, the guided span maximises the local
boundary-weighted score sensitivity. The two main simulations then test whether
this allocation improves the final independently calibrated region.

At active rank $r=4$, with the anchor, score Jacobian, curvature metric,
main-chain budget, completion, guard and calibration procedure matched across
constructions, normalized disagreement with the population global-level oracle
region is $\numSimOneGuidedErr\%$ versus $\numSimOneUniformErr\%$ for
uniform weighting on Simulation~1 and $\numSimTwoGuidedErrRfour\%$ versus
$\numSimTwoUniformErrRfour\%$ on Simulation~2. These correspond to reductions
of about $22\%$ in both simulations. The multiplicity-adjusted paired 95\%
confidence intervals for guided minus uniform are
$[\numSimOneDiffLo,\numSimOneDiffHi]$ and
$[\numSimTwoDiffLo,\numSimTwoDiffHi]$ percentage points, respectively, and
exclude zero in both cases (Table~\ref{tab:bwc-main}). Appendix~\ref{app:sim-guidance-r4}
gives the matched design and region-error definition, while
Appendix~\ref{app:bwc-combined} contains additional mechanism controls.

The matched rank sweep shows the largest benefit at low rank, with the
advantage statistically unresolved at rank $32$ on both simulations
(Table~\ref{tab:guidance-rank-sweep}). These fixed-anchor comparisons support
direction selection at limited rank; its contribution under the deployed
adaptive-rank and anchor-selection rules remains to be isolated.

\subsection{Predictive Accuracy and Computational Efficiency}
\label{sec:fidelity}
\label{sec:efficiency}

Relative to the specified HMC baseline, \SCOPE's signed test-NLL difference
has median $\numScopeNllGapMedian$ nats and is at most $0.1$ on
$\numNScopeWithinPointOne$ of $\numNMain$ benchmarks. The largest excesses
are $0.193$ on Concrete and $\numScopeNllGapWorst$ on Superconductivity
(paired intervals: Table~\ref{tab:app-paired}). Median inference time after
shared initialisation is $\numScopeTimeMedianEtoE$ seconds versus
$\numHmcTimeMedian$ seconds on the six UCI datasets. Full MALA is the closest
low-cost competitor in median NLL, with worst-benchmark excess
$\numMalaNllGapWorst$ nats (Table~\ref{tab:app-nll}).
Across the six UCI datasets, the median ratio of \SCOPE's predictive ESS/s to
HMC's is $\numPessScopeMedian\times$, or $\numPessChainMedian\times$ using
active-chain states alone; both charge SCOPE's full reported inference time.
For the slowest-mixing test point, the corresponding ratios are
$\numPessWorstMedian\times$ and $\numPessChainWorstMedian\times$.
Full MALA gives $\numPessMalaMedian\times$ and
$\numPessMalaWorstMedian\times$ on the median- and slowest-point summaries.
These ESS comparisons describe autocorrelation within each method's
approximation; predictive quality is assessed separately by scores and region
metrics. Table~\ref{tab:app-predess} separates active-chain and completed draws.

\begin{table}[t]
\centering
\caption{\textbf{Bayesian predictive comparison (20 matched seeds).}
Predictive summaries use the eight main benchmarks; runtime and predictive
ESS/s use the six main UCI datasets. SCOPE includes both candidate fits and
converged-MAP construction; baselines charge a single selected fit. NLL excess
is test NLL minus HMC. ESS/s is for log predictive density relative to HMC,
reported at the median and slowest-mixing test point and then summarised across
the six UCI datasets; parentheses give the active MCMC chain of SCOPE without
completion. Lower-cost methods take at most a tenth of HMC's median wall-clock.
Bold marks the best lower-cost summary, not statistical significance. Full
per-benchmark values and region comparisons are in the appendix.}
\label{tab:main}
\tiny
\setlength{\tabcolsep}{5pt}
\renewcommand{\arraystretch}{1.05}

\resizebox{\textwidth}{!}{%
\begin{tabular}{@{}l rr rr rr@{}}
\toprule
& \multicolumn{2}{c}{Wall-clock / seed}
& \multicolumn{2}{c}{NLL $-$ HMC ($\downarrow$)}
& \multicolumn{2}{c}{Predictive ESS/s vs HMC ($\uparrow$)}\\
\cmidrule(lr){2-3}
\cmidrule(lr){4-5}
\cmidrule(lr){6-7}

Method
& s
& vs HMC
& median
& worst
& median point
& slowest point\\

\midrule
\multicolumn{7}{@{}l}{\emph{Lower-cost methods}}\\[1pt]

Laplace
& \textbf{71}
& \textbf{17$\times$}
& +0.062
& +0.938
& iid
& iid\\

Trajectory Gaussian
& 91
& 13$\times$
& +0.087
& +0.986
& iid
& iid\\

MFVI
& 94
& 12$\times$
& +0.176
& +0.585
& iid
& iid\\

Trajectory PCA
& 92
& 12$\times$
& +0.210
& +4.152
& 1.7$\times$
& 2.9$\times$\\

SGLD
& 105
& 11$\times$
& +0.381
& +1.834
& 3.5$\times$
& 15.4$\times$\\

Full MALA
& 106
& 11$\times$
& +0.011
& +0.771
& 0.85$\times$
& 1.3$\times$\\

\rowcolor{scoperow}
\textbf{SCOPE}
& 89
& 13$\times$
& \textbf{+0.002}
& \textbf{+0.266}
& \textbf{30$\times$} (20$\times$)
& \textbf{106$\times$} (72$\times$)\\

\midrule
\multicolumn{7}{@{}l}{\emph{HMC-cost references}}\\[1pt]

Deep ensemble
& 1{,}634
& 0.8$\times$
& +0.032
& +1.294
& --
& --\\

Full HMC
& 1{,}171
& 1$\times$
& 0
& 0
& 1$\times$
& 1$\times$\\

\bottomrule
\end{tabular}}
\end{table}

\subsection{Calibrated Regions and Predictive Geometry}
\label{sec:regions}
\label{sec:sim2}

The guide-selected level of \SCOPE\ gives realised coverage
$\numScopeCovLo$--$\numScopeCovHi$ across the $\numNMain$ main benchmarks,
while independent calibration of the same predictive (\SCOPE/Split) gives
$\numSplitCovLo$--$\numSplitCovHi$. The two choices differ in coverage by at
most $\numReuseGapMaxAbs$ and in region size by a factor of
$\numReuseSizeRatioLo$--$\numReuseSizeRatioHi$ (Appendix
Table~\ref{tab:app-split}); these empirical differences do not provide a numerical estimate of the complexity penalty in Theorem~\ref{thm:reuse}. Among methods with the same
finite-sample marginal guarantee, realised coverage is uniformly close to
$0.90$, so size and geometry are the informative comparisons. See Appendix Table~\ref{tab:app-finite-sample-summary} for the complete
comparison.

Simulation~2 illustrates the flexibility of density-level regions:
Figure~\ref{fig:sim2} shows a disconnected SCOPE region, while CQR spans the
low-density gap. This illustrates region representation; the matched weighting
experiments isolate score guidance. Against the population oracle,
independently calibrated \SCOPE/Split has mean symmetric-difference length
$0.83$, versus $\numSymDiffSimTwoCqr$ for CQR and $0.81$ for HMC+Conf
(Table~\ref{tab:app-finite-sample-summary}). Predictive TV distance to the
known conditional law is $\numTvTrueSimTwoScope$ for SCOPE and
$\numTvTrueSimTwoHmc$ for HMC (Table~\ref{tab:app-sims}).

\subsection{Completion and Guide Reuse (H2, H3)}
\label{sec:h2h3}

\ifCompletionReady
Guarded completion improves mean NLL over the same active chain alone on all
$\numHTwoN$ benchmarks studied, with coverage changes at most
$\numHTwoCovDiffMaxAbs$. The guard materially affects this result:
unguarded completion increases NLL relative to guarded completion by up to
$\numHTwoUnguardNllPenaltyMax$ nats (Appendix~\ref{app:completion}).
\fi
\ifGuideSizeReady
When the complete guide-dependent procedure is rerun at each guide size on the
four-dataset study, the coverage difference between \SCOPE\ and \SCOPE/Split
ranges from $\numHThreeGapMin$ to $+\numHThreeGapMax$. This describes observed finite-sample behaviour; it does not evaluate the complexity term in Theorem~\ref{thm:reuse} or provide a numerical certificate for the guide-reuse bound (Appendix~\ref{app:guide-size}).
\fi

\section{Discussion and Conclusion}
\label{sec:discussion}

\SCOPE\ uses boundary-weighted score gradients to allocate posterior
computation to a selected subspace, then calibrates regions from the resulting
approximate predictive. The local sensitivity result motivates this allocation;
matched simulation experiments show reduced oracle-region error at low fixed
rank. The default pipeline combines score guidance, anchor selection, slice
MCMC and guarded variational completion. Its benchmark results show a
predictive-score--inference-time tradeoff relative to the specified HMC
baseline, with near-nominal empirical coverage. These comparisons assess the
complete construction; the component controls isolate direction selection
under their matched settings. The inference-time comparisons are relative
to the specified fixed-budget HMC configuration, not a universal speedup
over all HMC implementations or tuning choices. Independent calibration
gives \SCOPE/Split finite-sample marginal validity, while guide reuse has the
complexity-dependent guarantee of Theorem~\ref{thm:reuse}.

\SCOPE\ also has limitations. The active chain targets the posterior restricted
to an affine slice rather than the exact marginal distribution of the active
coordinates, and the remaining uncertainty is represented by a diagonal
variational approximation followed by an energy guard. The score-guided
geometry is local and coverage-level dependent, so changing the target coverage
level requires rebuilding the subspace. In addition, the guide-reuse guarantee depends on the complexity of a guide-independent class containing the fitted region rule. Appendix
~\ref{sec:scope-complexity} gives asymptotic decay for the fixed configuration
considered here, although the resulting finite-sample bound can be vacuous at
the sample sizes studied. \SCOPE/Split avoids this issue by using an
independent calibration set and therefore has ordinary finite-sample marginal
validity under exchangeability
(Proposition~\ref{prop:split-validity}).

Natural directions for future work include extending \SCOPE\ to substantially
larger models, structured outputs, developing richer
representations of residual uncertainty, and constructing score-guided
subspaces that can be shared across coverage levels. For large pretrained
models, score guidance could select directions within low-rank adapters,
which already make Bayesian fine-tuning tractable
\citep{yang2024bayesian}. Extensions to multivariate responses could use
joint-density scores, complementing existing approaches to multivariate
conformal prediction \citep{feldman2023calibrated}, while structured outputs
would require suitable scores and tractable prediction-set constructions.
Under distribution shift, score guidance could be adapted alongside weighted
conformal calibration for covariate shift \citep{tibshirani2019conformal}
or adaptive conformal calibration for sequential data
\citep{gibbs2021adaptive}. Another direction is to investigate Bayesian models
over active subspaces, using inference on matrix manifolds to represent
uncertainty in their orientation \citep{holbrook2016matrix}. When likelihood
or gradient evaluations are expensive, as in Bayesian inverse problems and
large Bayesian neural networks, the active chain could be combined with
neural-network emulators of the likelihood
\citep{lan2022scaling,moslemi2024scaling}. More broadly, our results suggest
that the downstream object ultimately reported by a Bayesian analysis can
help determine where approximate inference should spend its computational
effort.

\FloatBarrier

\appendix
\section{Proofs}

\subsection{Proof of Proposition~\ref{prop:local-relevance}}
\label{app:prop1}

For guide observation $i$, first-order Taylor expansion at $\thetastar$ gives
\[
\rho_{\thetastar+tH_\star^{-1/2}v}(x_i^g,y_i^g)
-
\rho_{\thetastar}(x_i^g,y_i^g)
=
t(J_\star H_\star^{-1/2}v)_i+o(t).
\]
Squaring, weighting, and summing yields
\begin{multline*}
\sum_i
\omega_i
[
\rho_i(\thetastar+tH_\star^{-1/2}v)-\rho_i(\thetastar)
]^2
\\
=
t^2
v^\top
H_\star^{-1/2}
J_\star^\top
\operatorname{diag}(\omega)
J_\star
H_\star^{-1/2}
v
+
o(t^2),
\end{multline*}
which is \eqref{eq:firstorder}. If
$J_\star H_\star^{-1/2}v=0$, the leading term vanishes.

For a fixed local score level $c$, suppose a scalar boundary point
$y_c(x,\theta)$ satisfies
$\rho_\theta(x,y_c)=c$ and
$\partial_y\rho_{\thetastar}(x,y_c)\neq0$.
The implicit-function theorem gives
\[
\frac{d y_c}{dt}\bigg|_{t=0}
=
-
\frac{
\nabla_\theta\rho_{\thetastar}(x,y_c)^\top
H_\star^{-1/2}v
}{
\partial_y\rho_{\thetastar}(x,y_c)
}.
\]
This identity relates score sensitivity at a boundary point to boundary motion
for a fixed level. A changing level contributes its own derivative, so
Proposition~\ref{prop:local-relevance} does not characterise the motion of the
recalibrated region. The membership pilot recomputes state-specific quantiles
but refines only the observation weights in the score operator.

\paragraph{Rank-$r$ statement and spectral-mass rule.}
Let
$A=J_\star^\top\operatorname{diag}(\omega)J_\star$ and
$M=M_\star(\omega)=H_\star^{-1/2}AH_\star^{-1/2}$.
For any $B$ satisfying $B^\top H_\star B=I_r$, write
$U=H_\star^{1/2}B$. Then
\[
\operatorname{tr}(B^\top AB)
=
\operatorname{tr}(U^\top MU),
\qquad
U^\top U=I_r.
\]
By the Ky Fan maximum principle,
\[
\max_{U^\top U=I_r}
\operatorname{tr}(U^\top MU)
=
\sum_{j=1}^{r}\lambda_j,
\]
attained at $U=P_r$, so $B_r=H_\star^{-1/2}P_r$. Hence no other
rank-$r$ curvature-orthonormal subspace captures more total local weighted
score sensitivity. If $r_\eta$ is the smallest rank satisfying
$\sum_{j=1}^{r_\eta}\lambda_j\ge\eta\sum_j\lambda_j$, then for any
$r<r_\eta$ the Ky Fan bound is strictly below $\eta\sum_j\lambda_j$. This
proves the rank-$r$ and minimum-rank statements in
Proposition~\ref{prop:local-relevance}.
\hfill$\square$

\paragraph{Dual construction.}
For the dual construction, let
\[
K(\omega)
=
\operatorname{diag}(\omega)^{1/2}
J_\star H_\star^{-1/2}.
\]
Then
$M=K^\top K$ and $G=KK^\top$.
If
$Gv_j=\lambda_jv_j$ with $\lambda_j>0$, then
\[
p_j
=
\frac{K^\top v_j}{\sqrt{\lambda_j}}
\]
is the corresponding normalised eigenvector of $M$.

\subsection{Proof of Proposition~\ref{prop:gaussian}}

Since
$C_B=B^\top\SigVI^{-1}B$,
\[
\operatorname{Cov}(Ba_G)
=
BC_B^{-1}B^\top.
\]
By independence,
\[
\operatorname{Cov}(Ba_G+\delta)
=
BC_B^{-1}B^\top
+
\SigVI
-
BC_B^{-1}B^\top
=
\SigVI.
\]
The sum is Gaussian and centred, so
\[
\thetastar+Ba_G+\delta
\sim
\N(\thetastar,\SigVI).
\]
When $\thetastar=\thetaVI$, this coincides with the fitted mean-field
variational reference; otherwise it is an anchor-centred Gaussian with the
same fitted residual scales.
\hfill$\square$

\subsection{Proof of Proposition~\ref{prop:split-validity}}
Condition on everything used to construct $\widehat p$. By assumption, the
$m_c$ calibration scores and the future score are exchangeable. The rank of
the future score among these $m_c+1$ scores is therefore uniform up to ties;
using the conservative order statistic defined in Section~\ref{sec:validity} gives
coverage at least $1-\alpha$ by the standard split-conformal argument.
\hfill$\square$

\subsection{Proof of Theorem~\ref{thm:reuse}}

Work under the conditioning in the theorem. Let $P$ be the common law of
the observations and write $Z_i^g=(x_i^g,y_i^g)$,
$Pg=\int g\,dP$, and
$P_{m_g}g=m_g^{-1}\sum_{i=1}^{m_g}g(Z_i^g)$.
Define the fitted region-membership function
\[
g_{\Dguide}(x,y)
=\mathbf 1\{-\log\pSCOPE(y\mid x;\Dguide)\le\widehat q_g\}.
\]
By construction, $g_{\Dguide}\in\mathcal G_{m_g}$ almost surely.
The threshold rule in Section~\ref{sec:level} gives
$P_{m_g}g_{\Dguide}\ge k_g/m_g\ge1-\alpha$ when $k_g\le m_g$,
including any ties. When $k_g=m_g+1$, the threshold is $+\infty$
and $g_{\Dguide}\equiv1$. Hence, in either case,
\[
Pg_{\Dguide}
\ge P_{m_g}g_{\Dguide}
-\sup_{g\in\mathcal G_{m_g}}(P_{m_g}g-Pg)
\ge1-\alpha-\sup_{g\in\mathcal G_{m_g}}(P_{m_g}g-Pg).
\]

Let $\Dguide'=\{Z_i^{g\prime}\}_{i=1}^{m_g}$ be an independent copy of
the guide sample, and let
$P_{m_g}'g=m_g^{-1}\sum_{i=1}^{m_g}g(Z_i^{g\prime})$.
The class $\mathcal G_{m_g}$ is fixed independently of both samples.
Taking the supremum outside the expectation over $\Dguide'$ and then
symmetrizing gives
\[
\begin{aligned}
\E_{\Dguide}\sup_{g\in\mathcal G_{m_g}}(P_{m_g}g-Pg)
&=\E_{\Dguide}\sup_{g\in\mathcal G_{m_g}}
  \E_{\Dguide'}[P_{m_g}g-P_{m_g}'g\mid\Dguide]\\
&\le\E_{\Dguide,\Dguide'}\sup_{g\in\mathcal G_{m_g}}
  \frac1{m_g}\sum_{i=1}^{m_g}
  \bigl[g(Z_i^g)-g(Z_i^{g\prime})\bigr]\\
&=\E_{\Dguide,\Dguide',\sigma}\sup_{g\in\mathcal G_{m_g}}
  \frac1{m_g}\sum_{i=1}^{m_g}\sigma_i
  \bigl[g(Z_i^g)-g(Z_i^{g\prime})\bigr]\\
&\le\E_{\Dguide,\sigma}\sup_{g\in\mathcal G_{m_g}}
  \frac1{m_g}\sum_{i=1}^{m_g}\sigma_i g(Z_i^g)\\
&\quad+\E_{\Dguide',\sigma}\sup_{g\in\mathcal G_{m_g}}
  \frac1{m_g}\sum_{i=1}^{m_g}(-\sigma_i)g(Z_i^{g\prime})\\
&=2\mathfrak R_{m_g}(\mathcal G_{m_g}).
\end{aligned}
\]
The second equality follows because each pair $(Z_i^g,Z_i^{g\prime})$
is exchangeable and the pairs are independent, so randomly swapping their
members leaves their joint law unchanged. The last equality uses the
symmetry of the Rademacher signs.

Independence of the future observation from $\Dguide$ gives
\[
\Prb\{Y_{\mathrm{new}}\in\CSCOPE(X_{\mathrm{new}})\mid\Dguide\}
=Pg_{\Dguide}.
\]
Taking expectations over $\Dguide$ therefore yields
\[
\Prb\{Y_{\mathrm{new}}\in\CSCOPE(X_{\mathrm{new}})\}
=\E_{\Dguide}[Pg_{\Dguide}]
\ge1-\alpha-2\mathfrak R_{m_g}(\mathcal G_{m_g}).
\]

\subsection{Complexity of the SCOPE Region Class}
\label{sec:scope-complexity}

Theorem~\ref{thm:reuse} bounds the marginal coverage loss from guide reuse by
$2\mathfrak R_{m_g}(\mathcal G_{m_g})$, where $\mathcal G_{m_g}$ is a
guide-independent class containing the fitted region-membership function.
Here we construct such a class for \SCOPE, identify the sources of its
complexity, and give a conservative growth bound. Throughout, condition on
$\Dtrain$ and data-independent simulation randomness, and retain the
measurability assumptions of the theorem. Let $\mathcal Z$ be the observation
space, $D$ the parameter dimension, and $N_s$ the fixed number of retained
main-chain states per candidate.

\paragraph{Why a guide-independent class is legitimate.}
Write $\mathcal A_{m_g}(d)$ for the score rule returned by the complete fitting
procedure on a possible guide sample $d$ of size $m_g$. Its range
\[
\mathcal R_{m_g}^{\mathrm{out}}
=\{\mathcal A_{m_g}(d):d\in\mathcal Z^{m_g},\ \mathcal A_{m_g}(d)\text{ is defined}\}
\]
is fixed before the actual $\Dguide$ is observed, although
$\mathcal A_{m_g}(\Dguide)$ is data-dependent. Thus class independence is not
an assumption that \SCOPE\ ignores the guide labels. The substantive
question is whether this range, or a measurable containing class, is small
enough to generalize. Taking all measurable score rules would satisfy
containment but give no useful guarantee.

\paragraph{An explicit containing family.}
Let $\mathcal I\subseteq\{1,\ldots,D\}$ be the fixed set of allowed ranks
and $a\in\{\mathrm{VI},\mathrm{MAP}\}$ index the two training-fitted anchors.
Write $H_a\succ0$ for their curvature metrics and $\Sigma=\SigVI\succ0$
for the training-fitted VI covariance. For the argument, index independent
raw completion draws $\epsilon_{a,s}\sim\N(0,\Sigma)$ by candidate and
retained-state number before observing the guide sample. Generating these
independent draws in advance leaves the sampling law unchanged; they are
fixed under our conditioning.

For each $a,r$, allow any
\[
\begin{aligned}
B&\in\mathcal B_{a,r}:=\{B\in\R^{D\times r}:B^\top H_aB=I_r\},\\
A&=(a_1,\ldots,a_{N_s})\in\R^{r\times N_s},\\
c&=(c_1,\ldots,c_{N_s})\in[0,1]^{N_s}.
\end{aligned}
\]
Define
\begin{align}
\Pi_B&=I_D-B(B^\top\Sigma^{-1}B)^{-1}B^\top\Sigma^{-1},
\nonumber\\
\vartheta_{s,\pm}&=\theta_a+Ba_s\pm c_s\Pi_B\epsilon_{a,s},
\nonumber\\
\widetilde p_{a,r,B,A,c}(y\mid x)
&=\frac1{2N_s}\sum_{s=1}^{N_s}
\left[p_{\vartheta_{s,+}}(y\mid x)+p_{\vartheta_{s,-}}(y\mid x)\right].
\label{eq:complexity-envelope}
\end{align}
The matrix inverse exists because $B$ has full column rank. Let
$\overline{\mathcal R}$ consist of the negative log densities in
\eqref{eq:complexity-envelope}; the positive Gaussian densities used here
give real-valued scores. Its membership class is
$\overline{\mathcal G}=\bigcup_{a,r}\mathcal G_{a,r}$, where
\[
\begin{aligned}
\mathcal G_{a,r}
=\bigl\{&(x,y)\mapsto
\mathbf 1\{\widetilde p_{a,r,B,A,c}(y\mid x)\ge u\}:\\
&B\in\mathcal B_{a,r},\ A\in\R^{r\times N_s},\
c\in[0,1]^{N_s},\ u\ge0\bigr\}.
\end{aligned}
\]
The level $u=e^{-t}$ also includes $u=0$ when $t=+\infty$.
Every \SCOPE\ output belongs to this family: substitute its selected anchor
and rank, final basis, retained active states and guard scales. The family
is deliberately larger than the algorithm's range, because arbitrary
$A$ and $c$ need not be attainable by MALA and the guard. Nevertheless,
$\overline{\mathcal R}$ is a valid choice in Theorem~\ref{thm:reuse} and
provides an upper bound for any smaller containing class.

\paragraph{Where the complexity enters.}
\emph{Training and anchor selection.}
Training fixes $\theta_a$, $H_a$, $\Sigma$, the network architecture and the
posterior target. These are not additional quantities fitted to the guide
sample. Selecting between two anchors takes a union of two families,
rather than selecting between two fixed score rules, since each candidate
has its own guide-dependent geometry and predictive.

\emph{Boundary weighting and rank selection.}
The guide gradients, empirical boundary level, MAD bandwidth and spectral
rule select $B$ and $r$. Even for fixed $r$, an arbitrary $r$-dimensional
span in $\R^D$ has $r(D-r)$ degrees of freedom. Thus the dimension $r$ of
the sampled coordinates does not measure the flexibility of choosing their
span. The constraint $B^\top H_aB=I_r$ restricts the family, but the realized
basis cannot be treated as fixed independently of the guide set.
In the conservative count below we retain all $Dr$ entries of $B$, rather
than exploit its constraints.

\emph{Membership refinement.}
The pilot updates the weights and hence the final basis. It does not
introduce an additional set of free guide weights into
\eqref{eq:complexity-envelope}: all its effects on the geometry are already
covered by allowing $B$ to vary, and any effects on the retained states are
covered by $A$. A smaller class could exploit which bases the weighted
score operator can actually produce. For example, its positive-eigenvalue
directions lie in the fixed space
\[
\mathcal V_a
=\operatorname{span}\{H_a^{-1/2}\nabla_\theta\rho_{\theta_a}(z):
z\in\mathcal Z\}.
\]
If this space has dimension $d_a<D$, the corresponding whitened bases can
be parameterized with $d_ar$ entries in place of $Dr$. This reduction
requires that only positive-eigenvalue directions are selected; any
directions added in a degenerate case must also be included. In contrast,
the span of the gradients on the observed guide sample is data-dependent
and cannot simply be frozen for the bound.

\emph{Active MCMC.}
The $N_s$ retained vectors contribute at most $N_sr$ free coordinates to
the containing family. Their Markov dependence does not invalidate the
coverage argument, which requires independent guide observations, not
independent parameter draws. This envelope also makes no mixing or
convergence assumption about the chain. Those properties matter for
posterior approximation, whereas containment is enough for this coverage
bound. Increasing $N_s$ enlarges this conservative parameter count; it
does not establish that longer chains worsen actual coverage.

\emph{Variational completion and the guard.}
The raw $\epsilon_{a,s}$ are fixed under conditioning, and their projections
are determined by $B$ and the training-fitted $\Sigma$. They do not supply
$N_s(D-r)$ additional freely adjustable residual coordinates. Allowing
one shrinkage factor $c_s$ per retained state covers the guide-dependent
guarded outputs. The antithetic pair shares this factor and the same active
state. If the VI covariance were also refitted using guide data, the
containing family would have to include that additional variation.

\emph{Final calibration.}
The density level $u$ contributes one scalar. For a single fixed score
rule, thresholding gives nested sets and at most $k+1$ membership patterns
on $k$ observations. The larger complexity comes from varying the fitted
score rule as well. Because thresholding is discontinuous, a bound for raw
scores or densities cannot automatically be substituted for a bound on
their membership class.

\paragraph{A conservative quantitative bound.}
For a binary class $\mathcal G$, define its growth function by
\[
\mathsf S_{\mathcal G}(k)
=\sup_{z_1,\ldots,z_k}
\left|\{(g(z_1),\ldots,g(z_k)):g\in\mathcal G\}\right|.
\]
For fixed $a,r$, the family above has the real parameter count
\begin{equation}
W_r=\underbrace{Dr}_{B}
+\underbrace{N_sr}_{A}
+\underbrace{N_s}_{c}
+\underbrace{1}_{u}.
\label{eq:complexity-count}
\end{equation}
Parameter counting alone is not a VC-dimension bound. For the fitted
Gaussian neural models, however, the required regularity can be supplied
by the form of their operations. Assume a fixed architecture built from
affine maps, ReLU, softplus, sigmoid or $\tanh$, with positive Gaussian
scales and a fixed number of likelihood-mixture components. These
operations, the projection in \eqref{eq:complexity-envelope}, and the
density comparison are definable in the real exponential field, an
o-minimal structure \citep[Theorem~5.2]{wilkie2009ominimal}. In this setting,
the growth of a family indexed by $W_r$ real parameters is bounded by
$C_{a,r}k^{W_r}$ \citep[Theorem~1.1]{aschenbrenner2016vc}.
Here \emph{definable} means expressible using finitely many arithmetic
operations, comparisons, exponentials, and logical quantifiers; logarithms
are included as inverses of exponentials. The cited result supplies the
growth bound, not an estimate of its constant. It applies to the fitted
model, without requiring the data-generating distribution to belong to
that model.

Let $K=2|\mathcal I|$, $W=\max_{r\in\mathcal I}W_r$, and choose a finite
$C\ge1$ bounding all the $C_{a,r}$. A union bound on membership patterns gives
\[
\mathsf S_{\overline{\mathcal G}}(k)
\le\sum_{a,r}\mathsf S_{\mathcal G_{a,r}}(k)
\le KCk^W.
\]
On any guide sample, there are at most
$\mathsf S_{\overline{\mathcal G}}(m_g)$ distinct binary vectors.
For such a vector $v$, the Rademacher moment bound
$\E_\sigma e^{\lambda\sum_i\sigma_i v_i}
\le e^{\lambda^2m_g/2}$ implies
\[
\E_\sigma\max_v\frac1{m_g}\sum_i\sigma_i v_i
\le \frac{\log|\{v\}|}{\lambda m_g}+\frac{\lambda}{2}.
\]
Optimizing over $\lambda>0$ and averaging over the guide sample yields
\begin{equation}
\mathfrak R_{m_g}(\overline{\mathcal G})
\le
\sqrt{\frac{2\log\mathsf S_{\overline{\mathcal G}}(m_g)}{m_g}}
\le
\sqrt{\frac{2\{\log(KC)+W\log m_g\}}{m_g}}.
\label{eq:complexity-growth}
\end{equation}
Consequently, Theorem~\ref{thm:reuse} gives coverage at least
$1-\alpha-2\sqrt{2\{\log(KC)+W\log m_g\}/m_g}$.
For fixed training data, architecture, allowed ranks, $N_s$ and simulation
randomness, this bound tends to $1-\alpha$ as $m_g\to\infty$.
It concerns the finite-mixture predictive actually used by \SCOPE, without
requiring convergence to an infinite-sample posterior predictive.

\paragraph{Interpretation and limitations.}
The analysis separates a finite choice among anchors and ranks, entering
through $\log K$, from continuous variation of the basis, active states,
completion and threshold. Its constants are not evaluated, and $W$ can
greatly exceed the available guide size. It therefore supplies a
fixed-configuration asymptotic justification, not a numerical certificate
for the reported coverage. If the architecture or retained-sample budget
grows with $m_g$, the corresponding constants and parameter counts must be
controlled again.

The containing family ignores the restrictions imposed by score guidance,
the MALA trajectory and the energy guard. Exploiting those restrictions,
or a smaller training-determined space $\mathcal V_a$, is a route to sharper
bounds. The bandwidth and pilot settings can change the reachable family,
but this worst-case envelope does not prove a monotone relation between
those settings and complexity. In particular, a large bandwidth leaves
the guide-dependent gradient matrix in place even when the initial weights
become nearly uniform. The weighted effective size
$1/\sum_i\omega_i^2$ is a geometry diagnostic, not a replacement for $m_g$
in the theorem, whose calibration uses unweighted guide scores.
Computational efficiency and statistical complexity are therefore related
questions that require separate evidence.

\section{Implementation Details}
\label{app:implementation}

\subsection{Notation}
\label{app:notation}

\begin{table}[htbp]
\centering
\caption{Core notation.}
\label{tab:notation}
\small
\setlength{\tabcolsep}{3pt}
\begin{tabularx}{\columnwidth}{@{}>{$}l<{$}X@{}}
\toprule
\text{Symbol} & \text{Meaning}\\
\midrule
\Dtrain,\Dguide,\Dconf,\Dtest & training, guide, independent calibration, test sets\\
n,\ m_g,\ m_c & sizes of $\Dtrain$, $\Dguide$, $\Dconf$\\
\rho_\theta(x,y) & density nonconformity score $-\log p_\theta(y\mid x)$\\
\thetaVI,\thetaMAP & candidate local references (VI mean, converged MAP)\\
\thetastar & guide-selected reference, eq.~\eqref{eq:anchor-selection}\\
\SigVI & diagonal VI covariance; residual scales\\
H_\star & positive-definite curvature metric at $\thetastar$ (Appendix~\ref{app:hstar})\\
J_\star & guide-score gradients at $\thetastar$ ($m_g\times D$)\\
\omega^{(0)},\omega^{(1)} & kernel and membership-refined guide weights\\
M_\star(\omega) & whitened score operator, eq.~\eqref{eq:operator}\\
B\in\R^{D\times r} & final active basis, eq.~\eqref{eq:basis}\\
a_s\in\R^r & retained active-coordinate state\\
\kappa,\ \kappa^\star & candidate and selected anchor indices
in $\{\mathrm{VI},\mathrm{MAP}\}$\\
N_0,N_s & numbers of retained pilot and main-chain states\\
\gamma_s & shrink-only energy-guard scale (Appendix~\ref{app:guard})\\
\pSCOPE & completed untempered SCOPE predictive, eq.~\eqref{eq:predictive}\\
\widehat q_g,\ \CSCOPE & reused-guide score level and region, eq.~\eqref{eq:main-region}\\
\mathfrak R_{m_g}(\mathcal G_{m_g}) & guide-reuse complexity term, Theorem~\ref{thm:reuse}\\
\bottomrule
\end{tabularx}
\end{table}

\subsection{Algorithm}
\label{app:algorithm}

\begin{algorithm}[t]
\caption{\SCOPE}
\label{alg:scope}
\small
\begin{algorithmic}[1]
\Require $\Dtrain,\Dguide,p_\theta,p_0,\alpha$; fixed optimiser, VI, rank and MALA settings
\Ensure untempered predictive $\pSCOPE$ and region $\CSCOPE$
\State Fit $\thetaVI,\SigVI$ and $\thetaMAP$ using $\Dtrain$.
\For{$a\in\{\mathrm{VI},\mathrm{MAP}\}$}
    \State Set $\thetastar=\theta_a$; compute guide scores, $q_0$, $H_\star$, and $\omega^{(0)}$~\eqref{eq:kernel}.
    \State Construct $B_0$ and run pilot MALA on its slice.
    \State Compute $u_i$, $\omega^{(1)}$~\eqref{eq:blend}, and the final basis $B$~\eqref{eq:basis}.
    \State Run main MALA targeting \eqref{eq:slice}; retain $a_1,\ldots,a_{N_s}$.
    \For{$s=1,\ldots,N_s$}
        \State Draw $\epsilon_s\sim\N(0,\SigVI)$; project and guard $\delta_s=\Pi_{\perp,\mathrm{VI}}\epsilon_s$.
        \State Set $\theta_{s,\pm}=\theta_a+Ba_s\pm\widetilde\delta_s$.
    \EndFor
    \State Form $\widehat p_a^{\,T=1}$~\eqref{eq:predictive} and record its guide NLL.
\EndFor
\State Select $a^\star$ by \eqref{eq:anchor-selection}; set $\thetastar=\theta_{a^\star}$ and freeze $\pSCOPE=\widehat p_{a^\star}^{\,T=1}$.
\State Compute unweighted guide scores under $\pSCOPE$ and their order statistic $\widehat q_g$.
\State \Return $\pSCOPE$, $\CSCOPE$~\eqref{eq:main-region}.
\end{algorithmic}
\end{algorithm}

\subsection{Converged MAP Construction}
\label{app:anchor}

The converged-MAP candidate uses only $\Dtrain$. Starting from the shared
staged MAP initialisation (Appendix~\ref{app:model}), we continue optimisation
with full-batch Adam at a constant learning rate (at most $4{,}000$ steps,
$\numAnchorAdamStepsLo$--$\numAnchorAdamStepsHi$ in practice) and $300$
L-BFGS iterations, which take $\numAnchorConvTimeLo$--$\numAnchorConvTimeHi$\,s,
retaining the best finite train log posterior encountered if the final
refinement does not improve it. The guide set is used only after both
candidate anchors have been constructed, through the predictive-NLL rule
\eqref{eq:anchor-selection}. Test diagnostics are never used to choose the
anchor in the main protocol.

\subsection{The Curvature Metric \texorpdfstring{$H_\star$}{H-star}}
\label{app:hstar}

For the heteroscedastic Gaussian likelihood, $H_\star$ is the Gauss--Newton
matrix obtained from Fisher information in the $(\mu,\sigma)$ outputs. With
$\nabla\mu_i$ and
$\nabla\sigma_i$ the gradients of $\mu_\theta(x_i)$ and $\sigma_\theta(x_i)$ at
$\thetastar$ for a training row $i$,
\begin{equation}
H_\star
=
\frac{n}{S}\sum_{i\in\mathcal S}
\Big[
\frac{\nabla\mu_i\nabla\mu_i^\top}{\sigma_i^2}
+
\frac{2\,\nabla\sigma_i\nabla\sigma_i^\top}{\sigma_i^2}
\Big]
+
\frac{1}{\sigma_p^2}I_D,
\label{eq:hstar}
\end{equation}
where $\mathcal S$ contains $S$ training rows drawn without replacement, with a
row budget of $512$. The same rows are used for every method requiring the
metric within a seed;
$\sigma_i$ is floored at $10^{-6}$, and $\sigma_p=0.25$ is the prior standard
deviation. The prior precision is the only damping in $H_\star$; this metric has
no additional ridge, trace-relative term or eigenvalue clipping. Simulation~2 instead uses
an empirical Fisher matrix for its mixture likelihood,
\[
\begin{aligned}
H_\star&=\frac{n}{S}\sum_{i\in\mathcal S}g_i g_i^\top+\sigma_p^{-2}I_D,\\
g_i&=\left.\nabla_\theta\log p_\theta(y_i\mid x_i)\right|_{\theta=\thetastar}.
\end{aligned}
\]
with $S=\min(n,2{,}048)$ rows drawn without replacement. Writing the data
term as $LL^\top$ ($L\in\R^{D\times2S}$ in the Gaussian case) and taking its
thin SVD
$L=V\Lambda^{1/2}U^\top$ (no truncation), $H_\star^{-1/2}$ is applied in closed
form,
\begin{equation}
H_\star^{-1/2}v
=
\sigma_p\,v
+
V\Big[\big(\Lambda+\sigma_p^{-2}I\big)^{-1/2}-\sigma_p I\Big]V^\top v,
\end{equation}
so no Cholesky factor or diagonal approximation of $H_\star$ is formed. The
score Jacobian $J_\star$ is computed at $\thetastar$ by reverse-mode
differentiation in standardised response units. The operator
$M_\star(\omega)$ of \eqref{eq:operator} is assembled explicitly as a
$D\times D$ matrix from the whitened, weighted rows of $J_\star$ and
eigendecomposed; eigenvalues are clamped at zero. Here $D$ ranges from $264$
to $4{,}933$ across the main benchmarks. The rank follows the $\eta=0.90$
spectral-mass rule with bounds
$[1,80]$. Separately, we report the spectral-tail diagnostic
\[
r_{\mathrm{cert}}
=
\min\left\{
r:
\big(\textstyle\sum_{j>r}\lambda_j\big)^{1/2}
\le
\varepsilon h
\right\},
\qquad \varepsilon=0.25,
\]
which asks whether the omitted first-order score variation is small relative
to the localisation bandwidth $h$; it is diagnostic only and never selects the
rank.

\subsection{Metropolis Correction on the Slice}
\label{app:mala-kernel}

For a fixed active basis $B$, let $q(a'\mid a)$ be the actual Langevin
proposal density, including any deterministic modification of its drift.
The acceptance probability is
\[
A(a,a')=\min\left\{1,\,
\frac{\exp\{\ell_n(\thetastar+Ba')\}\,q(a\mid a')}
     {\exp\{\ell_n(\thetastar+Ba)\}\,q(a'\mid a)}
\right\}.
\]
The forward and reverse densities use the same proposal convention,
including any drift clipping. Step-size adaptation is confined to discarded
warm-up; retained transitions use a fixed basis and fixed proposal settings.
This Metropolis kernel leaves the affine-slice target
\eqref{eq:slice} invariant.

\subsection{Shrink-Only Energy Guard}
\label{app:guard}

For a raw complement $\delta_s$, define
\[
D_s(c)
=
\ell_n(\theta_s^{\mathrm{act}})
-
\frac12
\left[
\ell_n(\theta_s^{\mathrm{act}}+c\delta_s)
+
\ell_n(\theta_s^{\mathrm{act}}-c\delta_s)
\right].
\]
For a Gaussian target and a matching Gaussian complement of dimension
$d_\perp=D-r$, the expected quadratic energy drop is $d_\perp/2$. Using this
as a reference, the implementation draws a target
$T_s\sim\tfrac12\chi^2_{d_\perp}$ and solves $D_s(\gamma_s)\approx T_s$ with
the local relation $D_s(c)\propto c^2$: starting from $c=1$, at most four
multiplicative updates $c\leftarrow c\sqrt{T_s/D_s(c)}$, clipped to
$[1/64,1]$, stopping when $|D_s(c)/T_s-1|\le0.10$. States whose ratio still
exceeds $1.3$, or is not finite, are halved up to three more times, and a
state that remains non-finite is pinned at $c=0$. Because $c\le1$ the guard
only ever shrinks the complement. The ``energy ratio'' diagnostic of
Table~\ref{tab:app-diag} is the within-fit median of
$D_s(1)/(d_\perp/2)$ on the raw complement, averaged over seeds;
``guard $\gamma$'' averages the within-fit median $\gamma_s$ in the same way.
The guard is a
numerical stabilisation, not a Metropolis correction, and
Proposition~\ref{prop:gaussian} describes the complement before it is applied;
the unguarded variant is reported in Appendix~\ref{app:completion}.
Across all eleven benchmarks, dataset means of the within-fit median guard
scale range from $\numGuardMedianLo$ to $\numGuardMedianHi$ and are below
$0.5$ on $\numGuardNBelowHalf$ benchmarks (Table~\ref{tab:app-diag}).

\subsection{Computational Cost}
\label{app:computation}

Each active-MALA proposal evaluates a full posterior gradient before
multiplication by $B^\top$, so low rank alone gives no $r/D$ cost reduction.
If $N_H$ HMC trajectories use $L$ leapfrog steps, the gradient budget is
approximately $G_{\mathrm{HMC}}\approx N_H(L+1)$. After initialisation, active
MALA needs approximately one new full gradient per proposal, giving
$N_{\mathrm{pilot}}+N_{\mathrm{main}}$ full gradient evaluations for the chains
of one SCOPE candidate. Both candidates are executed before selection, so the
complete procedure uses twice this chain budget, plus optimisation and geometry
work; the residual complement requires no inactive-space Markov chain.
Under the evaluation protocol detailed in Appendix~\ref{app:details},
HMC uses $\numHmcTrajectories$ trajectories with $L=\numHmcLeapfrog$,
approximately $\numHmcGradients$ full gradient evaluations per seed, whereas a
single SCOPE candidate uses $\numScopePilotProposals$ pilot and
$\numScopeMainProposals$ main proposals. Reported SCOPE inference time includes
both candidates' chains and geometry, converged-MAP construction, VI counted
once, completion, guarding, guide-NLL anchor selection and guide-score quantile
computation. It excludes the common initialisation, downstream metric evaluation
and region evaluation on the test grid (Appendix~\ref{app:metrics}).
Full-dimensional HMC must also use dynamics
stable for the stiffest directions of the full posterior, whereas the active
chain operates inside a curvature-scaled span selected for the downstream task.
The computational benefit depends on improved conditioning and the proposal
budget needed for predictive accuracy; it is not a formal $D/r$ complexity
result.

\subsection{Secondary Model-Mass Output}
\label{app:mhdr}

SCOPE/MHDR selects one predictive-spread scalar
by guide NLL and then reports a model-mass highest-density region
(Appendix~\ref{app:temperature}). This is predictive model selection, not
conformal calibration. The primary SCOPE region always uses the untempered
predictive; the spread search affects only the secondary SCOPE/MHDR output.
The reported primary runtime follows Table~\ref{tab:main} and the accounting
in Appendix~\ref{app:metrics}.

\section{Experimental Details}
\label{app:details}

\paragraph{Evaluation protocol.}
The main evaluation uses two simulations and six UCI regression datasets,
with Airfoil, Naval and Protein reported as additional benchmarks. Main
comparisons use $\numSeedCount$ matched seeds. The UCI datasets use
$50/20/15/15$ train/guide/calibration/test splits. All Bayesian methods share
the same network, data splits and MAP initialisation on $\Dtrain$. Bayesian
predictive comparisons include MFVI, diagonal Laplace, a trajectory Gaussian,
trajectory-PCA MCMC, a five-network deep ensemble, SGLD, full MALA and full
HMC. Finite-sample prediction-set comparisons use \SCOPE/Split together with
CQR, absolute- and normalised-residual conformal intervals, GBM-CQR and
split-conformal regions from selected Bayesian predictives. Full HMC is included as a specified fixed-budget, full-dimensional baseline
for predictive and computational comparison. On the simulations, accuracy
is additionally evaluated directly against the known data-generating law and
population oracle regions. The subsections below give the benchmark definitions, model and
sampler settings, metrics, runtime conventions and result-file provenance.

\subsection{Simulation 1}
\label{app:sim1}

The teacher is a sparse two-layer ReLU network in the same family as the fitted
mean network. Let
\begin{gather*}
X=(x,z,xz,u_1,\ldots,u_{12})^\top,\\
x\sim\operatorname{Unif}(0,1),\qquad
z\sim\operatorname{Bernoulli}(0.15),
\end{gather*}
and $U=(u_1,\ldots,u_{12})^\top\sim\N(0,\Sigma_X)$ with
$(\Sigma_X)_{jk}=0.3^{|j-k|}$. The eight active hidden units are
\[
\begin{aligned}
h_1&=\operatorname{ReLU}(x),& h_2&=\operatorname{ReLU}(1-x),\\
h_3&=\operatorname{ReLU}(z-0.5),& h_4&=\operatorname{ReLU}(xz),\\
h_5&=\operatorname{ReLU}(u_1),& h_6&=\operatorname{ReLU}(-u_1),\\
h_7&=\operatorname{ReLU}(u_2),& h_8&=\operatorname{ReLU}(-u_2).
\end{aligned}
\]
The second ReLU layer copies these nonnegative units, and
\[
\begin{aligned}
\mu_0(X)={}&-0.15+0.55h_1-0.25h_2+0.65h_3+0.55h_4\\
&+0.12h_5-0.12h_6-0.08h_7+0.08h_8,
\end{aligned}
\]
which on the support simplifies to
$-0.40+0.80x+0.325z+0.55xz+0.12u_1-0.08u_2$. The scale head is
\[
\sigma_0(X)=0.10+\operatorname{softplus}(1.25z+0.35xz-1.50).
\]
The rare subgroup is $\{z=1\}$ ($15\%$ of the population; $\numSeedCount$ seeds
give on average $176$ of the $1{,}200$ test points). Sample sizes are
$|\Dtrain|=1600$, $|\Dguide|=500$, $|\Dconf|=500$, $|\Dtest|=1200$.

\subsection{Simulation 2}
\label{app:sim2}

Simulation~2 is a stochastic two-component split--merge mixture with
$X\in[-2,2]^6$; only $x_1$ and $x_2$ enter the conditional law:
\begin{gather*}
m(x)=0.70\sin(1.30x_1)+0.35x_2,\\
d(x)=0.08+2.92\,\operatorname{sigmoid}\{2.40(x_1+0.45)\},\\
\mu_1(x)=m(x)-d(x),\qquad \mu_2(x)=m(x)+d(x),\\
\pi_2(x)=0.15+0.70\,\operatorname{sigmoid}(1.40x_2).
\end{gather*}
The fitted model is an ordered two-component Gaussian-mixture network, with
$|\Dtrain|=7{,}500$, $|\Dguide|=2{,}700$, $|\Dconf|=2{,}550$ and
$|\Dtest|=250$ in every seed. The designated subgroup is the stratum in which
the two components are separated (on average $182$ of the $250$ test points),
so subgroup coverage on this benchmark is a diagnostic of the disconnected
regime rather than a rare-group test.

\subsection{Real Benchmarks}
\label{app:realdata}

\begin{table}[htbp]
\centering
\caption{Real-data benchmarks. Splits are $50/20/15/15$ of the analysis sample
per seed, standardised on $\Dtrain$ only. The subgroup column gives the
registry's ``critical'' indicator; only predictor-defined indicators are used
as prediction-time subgroups in the main text.}
\label{tab:realdata}
\small
\setlength{\tabcolsep}{4pt}
\resizebox{\linewidth}{!}{%
\begin{tabular}{llrrrrrrl}
\toprule
Dataset & Role & Raw $n$ & Analysis $n$ & $d$ & Train & Guide & Conf & Subgroup indicator\\
\midrule
Concrete & main & 1030 & 1030 & 8 & 515 & 206 & 154 & age $>90$ days (predictor)\\
Energy & main & 768 & 768 & 8 & 384 & 154 & 115 & top-decile heating load (response)\\
Yacht & main & 308 & 308 & 6 & 154 & 62 & 46 & Froude number $>0.40$ (predictor)\\
Wine & main & 6497 & 6497 & 11 & 3248 & 1299 & 975 & quality $\ge7$ (response)\\
Abalone & main & 4177 & 4177 & 10 & 2088 & 835 & 627 & rings $\ge15$ (response)\\
Superconductivity & main & 21263 & 2000 & 81 & 1000 & 400 & 300 & $T_c>77$\,K (response)\\
\midrule
Airfoil & appendix & 1503 & 1503 & 5 & 752 & 301 & 225 & top-decile SPL (response)\\
Naval & appendix & 11934 & 2000 & 16 & 1000 & 400 & 300 & most-degraded decile (response)\\
Protein & appendix & 45730 & 2000 & 9 & 1000 & 400 & 300 & top-decile RMSD (response)\\
\bottomrule
\end{tabular}}
\end{table}

Table~\ref{tab:realdata} lists all nine UCI datasets: six main and three
additional datasets \citep{uci2023}. The test split is the
remaining $15\%$. Abalone's categorical sex variable is one-hot encoded
($d=10$); Wine pools the red and white subsets. For Superconductivity, Naval
and Protein a $2{,}000$-row analysis sample is drawn once with a fixed seed
before splitting so that the full-HMC reference remains affordable at the
$45{,}000$-trajectory budget.

\paragraph{Main and appendix benchmarks.}
The six main UCI datasets span analysis sample sizes from $308$ to $6{,}497$ and
$6$ to $81$ predictors, including low-noise targets (Yacht, Energy), noisier
targets (Wine, Abalone), and an ordinal target (Wine). Airfoil, Naval and
Protein are reported in full in Appendix~\ref{app:excluded}. These additional
tasks include nearly deterministic responses (Naval), a $4\%$ analysis
subsample (Protein), and mixed anchor selections (Airfoil:
$\numAnchorMapFracAirfoil/\numAnchorViFracAirfoil$ across seeds). Main-table
predictive and region aggregates use the eight main benchmarks (two simulations
and six UCI datasets); runtime and ESS/s aggregates use only the six main UCI
datasets. The full predictive tables cover eleven benchmarks (two simulations
and all nine UCI datasets).

\subsection{Model and Shared Initialisation}
\label{app:model}

The tabular Bayesian methods use the same heteroscedastic network: two hidden layers of
$16$ and $8$ ReLU units for the mean, a separate scale head with a softplus link
and a floor of $0.05$ standardised units, and an isotropic Gaussian prior with
standard deviation $0.25$ on all weights. The shared staged initialisation
uses mean pre-training for $100$ epochs, scale pre-training for $60$, and
joint warm-up for $30$, followed by an $80$-epoch MAP stage with Adam and
batch size $64$; it uses two restarts. The full shared initialisation
wall-clock is excluded from every timing. The converged-MAP candidate
continues this fit with full-batch
Adam (learning rate $2\times10^{-3}$, at most $4{,}000$ steps, patience $400$)
and $300$ L-BFGS iterations, keeping the best finite train log-posterior; it
uses $\Dtrain$ only.

\subsection{SCOPE Settings}
\label{app:hyper}

\begin{table}[htbp]
\centering
\caption{Default SCOPE settings used for every benchmark and seed.}
\label{tab:hyper}
\small
\setlength{\tabcolsep}{4pt}
\footnotesize
\begin{tabular}{@{}p{0.27\linewidth}p{0.30\linewidth}p{0.36\linewidth}@{}}
\toprule
Component & Hyperparameter & Default\\
\midrule
Candidate references & anchors & VI mean, converged MAP\\
Anchor choice & criterion & untempered guide NLL, eq.~\eqref{eq:anchor-selection}\\
Converged MAP & optimiser & full-batch Adam, $\le4000$ steps, then L-BFGS, 300 steps\\
VI & family / iterations / lr & diagonal Gaussian / $6000$ / $2\times10^{-3}$\\
Boundary localisation & $c_h$, $h_{\min}$ & $0.70$, $10^{-3}$\\
Fisher/GN metric & training-row budget & $512$ ($2{,}048$ on Sim.~2)\\
Operational rank & spectral mass $\eta$ / bounds & $0.90$ / $[1,80]$\\
Membership refinement & $\lambda_{\mathrm{blend}}$ & $0.50$\\
Pilot MALA & burn / retained / thin & $1000$ / $300$ / $3$\\
Main MALA & burn / retained / thin & $3000$ / $2000$ / $10$\\
MALA & target acceptance / initial step & $0.57$ / $0.05$\\
Completion & antithetic components & $K=2$\\
Energy guard & iterations / tolerance & $4$ / $0.10$\\
Guide level & calibration sample & all of $\Dguide$ (no internal split)\\
Response grid & points / padding & $500$ / $40\%$ of training range\\
SCOPE/MHDR only & spread scalar $c$ & guide-NLL line search\\
\bottomrule
\end{tabular}
\end{table}

Table~\ref{tab:hyper} lists the SCOPE defaults; none was tuned per benchmark.
The dataset-mean final active rank chosen by the spectral-mass rule ranges from
$\numRankLo$ to $\numRankHi$ across the main benchmarks (Simulation~2 hits the
cap of $80$); Appendix Table~\ref{tab:app-diag} reports per-benchmark ranks,
acceptance rates and the wall-clock breakdown of a fit.

\subsection{Baseline Settings}
\label{app:baselines}

Each fitted sampler candidate retains $2{,}000$ draws from one main chain
per seed; the selected candidate supplies the reported predictive.
\emph{Full HMC:} $5{,}000$ burn-in trajectories plus $2{,}000\times20$ thinned
trajectories of $20$ leapfrog steps, started at the shared MAP initialisation
with a small Gaussian jitter ($0.005$). The step size is tuned by dual
averaging towards acceptance $0.65$ (initial $0.025$; $\gamma=0.05$,
$t_0=10$, $\kappa=0.75$) during the first $80\%$ of burn-in and then frozen;
the mass matrix is the identity and is not adapted. Reported acceptance is
over all trajectories. This specification defines the fixed-budget HMC baseline used in the main
tables. Appendix~\ref{app:hmc-strong} reports an alternative
diagonally preconditioned, three-chain configuration as a sensitivity check.
\emph{Full MALA:} $10{,}000$ burn-in plus $2{,}000\times10$ transitions,
initialised at the guide-selected anchor, step size dual-averaged to acceptance
$0.57$ (initial $10^{-3}$, clipped to $[10^{-7},10^{-1}]$), gradient norm
clipped at $10^{4}$, identity preconditioner.
\emph{SGLD:} we use stochastic-gradient Langevin dynamics
\citep{welling2011sgld},
\[
\theta_{t+1}
=
\theta_t+\epsilon_t\,\widehat g_t
+\sqrt{2\epsilon_t}\,\xi_t,
\qquad
\xi_t\sim\N(0,I),
\]
where $\widehat g_t$ is an unbiased minibatch estimate of the full
log-posterior gradient. Minibatches contain $8$ observations drawn with
replacement, with the likelihood contribution rescaled by $n/8$.
\emph{MFVI:} diagonal Gaussian, $15{,}000$ Adam iterations (learning rate
$10^{-3}$), $2{,}000$ draws, initialised at the guide-selected anchor.
\emph{Laplace:} diagonal Hessian from $500$ Hutchinson probes, scale $0.5$,
$2{,}000$ draws; last-layer and subnetwork variants restrict the curvature to
the output layer and to the highest-variance subnetwork. \emph{Deep ensemble} \citep{lakshminarayanan2017ensembles}:
$M=5$ members, $7{,}000$ epochs each, mixture predictive. \emph{Trajectory
Gaussian (SWAG-style):} $600$ constant-learning-rate Adam steps
($2\times10^{-3}$) from the selected anchor, one snapshot every $10$ steps,
diagonal-plus-rank-$20$ covariance of the $60$ iterates. \emph{Trajectory-PCA
subspace MCMC:} the $120$ iterates of the same trajectory (one every $5$
steps), fixed rank $20$ with columns scaled by the trajectory standard
deviation, slice MALA from the trajectory mean with SCOPE's main-chain budget
and no complement; this is the \emph{baseline}. CQR: two-layer quantile network ($24$, $12$ units; $4{,}000$
epochs) conformalised on $\Dconf$; the point network for the split-conformal
interval and for normalised conformal has the same architecture as the
Bayesian model and is trained for $4{,}000$ epochs; GBM-CQR uses gradient
boosted quantile regression. FBNN/CES (functional emulator with $20$ latent
dimensions and pCN sampling) diverged on every real benchmark and is omitted.

\subsection{Metric Definitions}
\label{app:metrics}

\paragraph{Regions.}
Every region is represented on the common response grid (Table~\ref{tab:hyper}).
Coverage is the fraction of test responses inside the region; size is the
grid measure of the region in response units; both are averaged over seeds.
The split-conformal region of a density method uses the score
$R(x,y)=-\log\hat p(y\mid x)$ on $\Dconf$ with the usual
$\lceil(m_c+1)(1-\alpha)\rceil$-th order statistic; SCOPE's region uses the
same score and order statistic on $\Dguide$~\eqref{eq:main-region}; the
model-mass region is the highest-density set of $\hat p(\cdot\mid x)$ with
mass $0.90$ under the model itself.

\paragraph{Proper scores and discrepancies.}
On the simulations, TV and KL are computed against the known true conditional law.
On real datasets, TV and KL to the fixed-budget HMC predictive are reported
as agreement diagnostics rather than oracle-accuracy measures. On the simulations, the oracle prediction
region uses the known true conditional density at a single global density level
chosen for 90\% marginal coverage; symmetric difference is the grid measure of
the fitted region$\,\triangle\,$this population global-level oracle region.

\paragraph{Effective sample sizes.}
Within each recorded run, a $20$-dimensional task basis $Q_{\mathrm{task}}$
is fixed per seed (an
orthonormalisation of the $20$ leading directions of the boundary-weighted
guide-score operator~\eqref{eq:operator} at the shared MAP initialisation,
built with the same seed for every method) and used for its coordinate
diagnostics. The coordinate rate projects each method's $2{,}000$ retained draws on
$Q_{\mathrm{task}}$, computes the initial-positive-sequence ESS of each
coordinate (autocorrelations by FFT, consecutive lag pairs summed until the
first non-positive pair), and divides the 90th percentile over the $20$
coordinates by wall-clock; the 10th percentile and median are in
Appendix~\ref{app:ess-percentiles}. Task-mESS is estimated by batch means
(batch length $\lfloor\sqrt{n}\rfloor$). Main task-mESS summaries retain the
recorded estimates, which use the method's active basis when the common-basis
estimate is non-finite. The matched H1 table uses the common-basis summary
fields. In the
second run described under runtime accounting below the basis is built at the
converged MAP instead; the SCOPE, matched PCA, MALA-rerun and HMC-rerun rows
of that run share that basis, and wherever a ratio compares rows from the two
runs (a reference chain that was not rerun), the bases were constructed at
different anchors and need not have identical directions. ``Own-basis'' mESS
uses each method's own active basis and therefore measures mixing in a
different set of coordinates.
The two antithetic completion components are prediction components of the
same active state, not additional chain iterations, so SCOPE's sequence length
is also $2{,}000$. For the headline comparisons, a dataset's ESS/s is the
mean of the seedwise ESS/time rates. Table~\ref{tab:app-ess-percentiles}
instead divides mean ESS by mean inference time at each coordinate percentile;
these summaries need not coincide. Headline runtime ratios divide dataset-level mean times, and
headline ESS/s ratios divide dataset-level mean rates, before reporting
their median and range across benchmarks.
Paired comparisons instead use matched seed-level contrasts
(Appendix~\ref{app:paired}). The basis-free boundary and predictive-probe
functionals are defined in Appendix~\ref{app:functionals}.

\ifSpanSelReady
\paragraph{Selected-anchor span diagnostics.}
For the active-only span comparison, let $\mu_{si}$ and $\sigma_{si}$ be the
likelihood mean and standard deviation at test input $x_i$ for retained
state $s$. We report
\[
\rho=\frac{1}{n_t}\sum_{i=1}^{n_t}
\frac{\operatorname{sd}_s(\mu_{si})}{\operatorname{mean}_s(\sigma_{si})}.
\]
This is an average of local ratios to the likelihood scale, not a fraction
of total predictive variance. The test-side boundary diagnostic is
\[
\begin{aligned}
T^{\rm test}_{b,s}
&=\frac{1}{n_t}\sum_{i=1}^{n_t}
\operatorname{sigmoid}\!\left(\frac{q_{\rm ref}-R_s(x_i,y_i)}{0.2}\right),\\
b&=\operatorname{sd}_s(T^{\rm test}_{b,s}),
\end{aligned}
\]
where $R_s$ is the likelihood nonconformity score of state $s$ and
$q_{\rm ref}$ is the common frozen-reference level used for the chain
functionals. With completion, the smooth membership is averaged over the
antithetic components within each state before averaging over observations.
The standard deviation is taken after averaging test observations, so $b$
also depends on the covariance of their membership fluctuations. It is a
held-out functional-variability diagnostic, not an estimate of the error or
movement of the final calibrated prediction region.
\fi

\paragraph{Runtime accounting.}
A SCOPE candidate fit includes VI, active-geometry construction, the pilot
and main chains, completion and guarding. Reported SCOPE inference time
covers both candidate fits and converged-MAP construction, counting shared
VI once. Guide-NLL anchor selection and guide-score quantile computation are
included; the common initial MAP fit, downstream metric evaluation and region
evaluation on the test grid are excluded. Baseline times cover a single fit at the selected
anchor. Main-run timings used four CPU threads per seed on the same shared lab server. The frozen run metadata does not retain processor-model or memory fields, so we do not report hardware details that cannot be reconstructed.

Headline runtime and HMC-relative ESS/s comparisons use the six main UCI
benchmarks, for which the original run records both candidate costs.
Simulation inference timings were collected under different concurrent
machine loads. Their ESS/s ratios in Table~\ref{tab:app-predess} use
reconstructed SCOPE times and are descriptive estimates rather than paired
timing measurements; they are excluded from headline efficiency summaries.

\ifSplitReady
The final passes re-execute SCOPE with the same seeds and settings; the
simulation chains are re-drawn. The selected-anchor NLLs in
Table~\ref{tab:app-anchor} and completed NLLs in
Table~\ref{tab:app-completion} match the primary untempered SCOPE
predictive in Table~\ref{tab:app-nll}. Other controlled ablations use
the predictions and retained states of their stated comparison. The final-pass
UCI fit times are $\numFinalPassLoadLo$--$\numFinalPassLoadHi\times$ the
original times. Primary UCI timings use the original measurements.
Table~\ref{tab:app-functionals} reports diagnostics in final-pass seconds,
so those rates are compared within the final-pass run. Concurrent machine
load remains a source of timing variability.
\fi

\section{Additional Results and Ablations}
\label{app:ablations}

\subsection{Per-Benchmark Results}
\label{app:full-tables}

Table~\ref{tab:perdataset} reports SCOPE and HMC on the eight main benchmarks
(two simulations and six UCI datasets). Tables~\ref{tab:app-nll},
\ref{tab:app-confcov} and \ref{tab:app-confsize} add Airfoil, Naval and Protein,
covering all $\numNAll$ benchmarks (two simulations and nine UCI datasets). NLL refers
to the predictive distribution; coverage and size refer to the stated
region construction. The untempered SCOPE predictive defines the primary
region. Additional score, discrepancy and runtime matrices are retained in
the separate experimental results.

\begin{table}[htbp]\centering
\caption{\textbf{SCOPE versus full HMC: two simulations and six main UCI datasets (20 matched seeds).} SCOPE region: guide-reused level on the untempered predictive; HMC$+$Conf: the same density-score construction on the HMC draws, calibrated on $\Dconf$. Size ratio is SCOPE/HMC$+$Conf; $\Delta$NLL is SCOPE minus HMC (nats). Runtime and ESS/s comparisons use only the six main UCI datasets and the inference-time accounting of Table~\ref{tab:main}, including both SCOPE candidate fits after shared initialisation. The time ratio is HMC/SCOPE; ESS/s ratios are SCOPE/HMC. Ratios use unrounded dataset means. The recorded task-mESS estimates use an own-basis fallback when the common-basis estimate is non-finite; the matched H1 table instead uses common-basis summary fields (Appendix~\ref{app:metrics}). Simulation timing cells are omitted because final selected-anchor runs and frozen baselines had different execution loads.}
\label{tab:perdataset}
\small\setlength{\tabcolsep}{4pt}\renewcommand{\arraystretch}{1.05}
\resizebox{\linewidth}{!}{%
\begin{tabular}{l rr rr rr rr}\toprule
 & \multicolumn{2}{c}{Coverage} & & & \multicolumn{2}{c}{Wall-clock} & \multicolumn{2}{c}{ESS/s ratio}\\\cmidrule(lr){2-3}\cmidrule(lr){6-7}\cmidrule(lr){8-9}Benchmark & SCOPE & HMC & Size ratio & $\Delta$NLL & SCOPE (s) & HMC/SCOPE & Coord. & Task-mESS\\
\midrule
Sim.~1 & 0.902 & 0.894 & 1.03 & -0.008 & -- & -- & -- & --\\
Sim.~2 & 0.904 & 0.906 & 0.99 & -0.034 & -- & -- & -- & --\\
Concrete & 0.904 & 0.890 & 1.18 & +0.193 & 83 & 12$\times$ & 279$\times$ & 139$\times$\\
Energy & 0.918 & 0.895 & 1.17 & -0.071 & 77 & 15$\times$ & 416$\times$ & 188$\times$\\
Yacht & 0.907 & 0.898 & 1.00 & -0.211 & 71 & 14$\times$ & 52$\times$ & 52$\times$\\
Wine & 0.898 & 0.893 & 1.04 & +0.023 & 115 & 12$\times$ & 112$\times$ & 75$\times$\\
Abalone & 0.897 & 0.897 & 1.01 & +0.011 & 118 & 14$\times$ & 120$\times$ & 92$\times$\\
Supercond. & 0.906 & 0.896 & 1.29 & +0.266 & 95 & 13$\times$ & 149$\times$ & 94$\times$\\
\bottomrule
\end{tabular}}
\end{table}

\begin{table}[htbp]\centering
\caption{\textbf{Test NLL: two simulations and nine UCI datasets (11 benchmarks; means over 20 seeds; lower is better).} Bold: best in the column, full HMC included (the secondary SCOPE/MHDR output is excluded from the comparison); the shaded row is the SCOPE predictive behind the primary region. Simulation results use the anchor protocol specified in Appendix~\ref{app:details}.}
\label{tab:app-nll}
\small\setlength{\tabcolsep}{3pt}\renewcommand{\arraystretch}{1.05}
\resizebox{\linewidth}{!}{%
\begin{tabular}{lrrrrrrrrrrr}\toprule
Method & Sim.~1 & Sim.~2 & Concrete & Energy & Yacht & Wine & Abalone & Supercond. & Airfoil & Naval & Protein\\
\midrule
MFVI & $0.375$ & $1.453$ & $3.135$ & $1.152$ & $2.278$ & $1.053$ & $2.040$ & $3.843$ & $2.165$ & $-4.805$ & $3.008$\\
Laplace & $0.488$ & $1.376$ & $3.257$ & $0.887$ & $1.495$ & $1.069$ & $2.047$ & $4.491$ & $2.156$ & $-5.976$ & $3.000$\\
Trajectory Gaussian (Adam; SWAG-style) & $0.510$ & $1.920$ & $3.136$ & $0.839$ & $1.587$ & $1.067$ & $2.032$ & $4.539$ & $2.210$ & $-5.866$ & $2.990$\\
Trajectory-PCA subspace MCMC & $0.632$ & $2.363$ & $3.451$ & $0.845$ & $1.517$ & $1.192$ & $2.047$ & $7.705$ & $2.209$ & $-6.063$ & $3.140$\\
Deep ensemble & $0.473$ & $1.346$ & $3.055$ & $\mathbf{0.807}$ & $1.486$ & $1.035$ & $2.036$ & $4.847$ & $2.049$ & $-5.997$ & $3.009$\\
SGLD & $0.507$ & $1.718$ & $3.280$ & $1.832$ & $2.434$ & $1.114$ & $2.056$ & $5.387$ & $2.458$ & $-4.379$ & $3.253$\\
Full MALA & $0.377$ & $1.659$ & $3.049$ & $0.892$ & $1.683$ & $1.038$ & $2.031$ & $4.324$ & $2.087$ & $-6.072$ & $2.979$\\
Full HMC & $0.376$ & $1.293$ & $\mathbf{2.944}$ & $0.881$ & $1.693$ & $\mathbf{1.028}$ & $\mathbf{2.026}$ & $\mathbf{3.553}$ & $\mathbf{1.885}$ & $-6.051$ & $2.977$\\
Last-layer Laplace & $0.503$ & $1.816$ & $3.278$ & $0.841$ & $1.539$ & $1.078$ & $2.054$ & $5.355$ & $2.229$ & $-6.076$ & $3.043$\\
Subnetwork Laplace & $0.497$ & $1.802$ & $3.261$ & $0.846$ & $1.531$ & $1.078$ & $2.050$ & $5.072$ & $2.229$ & $-5.823$ & $3.021$\\
MAP density & $0.505$ & $2.016$ & $3.281$ & $0.839$ & $1.547$ & $1.147$ & $2.055$ & $5.382$ & $2.235$ & $-6.079$ & $3.256$\\
Frozen CABI (historical) & $0.417$ & $2.610$ & $3.156$ & $2.341$ & $2.619$ & $1.050$ & $2.112$ & $4.177$ & $2.271$ & $-3.112$ & $3.005$\\
SCOPE/MHDR (guide-tempered) & $0.376$ & $1.222$ & $3.127$ & $0.826$ & $1.423$ & $1.051$ & $2.038$ & $3.784$ & $1.995$ & $-6.534$ & $2.977$\\
\rowcolor{scoperow}\textbf{SCOPE (untempered predictive)} & $\mathbf{0.368}$ & $\mathbf{1.259}$ & $3.137$ & $0.810$ & $\mathbf{1.482}$ & $1.051$ & $2.037$ & $3.819$ & $1.909$ & $\mathbf{-6.954}$ & $\mathbf{2.969}$\\
\bottomrule
\end{tabular}}
\end{table}

\begin{table}[htbp]\centering
\caption{\textbf{Nominal-90\% coverage: two simulations and nine UCI datasets (means over 20 seeds).} SCOPE uses the guide-reused region; baselines use independent split calibration. Coverage is assessed relative to 0.90.}
\label{tab:app-confcov}
\small\setlength{\tabcolsep}{3pt}\renewcommand{\arraystretch}{1.05}
\resizebox{\linewidth}{!}{%
\begin{tabular}{lrrrrrrrrrrr}\toprule
Method & Sim.~1 & Sim.~2 & Concrete & Energy & Yacht & Wine & Abalone & Supercond. & Airfoil & Naval & Protein\\
\midrule
CQR & $0.900$ & $0.904$ & $0.893$ & $0.895$ & $0.913$ & $0.897$ & $0.898$ & $0.900$ & $0.904$ & $0.895$ & $0.894$\\
Abs.-residual conformal & $0.899$ & $0.904$ & $0.910$ & $0.903$ & $0.918$ & $0.895$ & $0.900$ & $0.903$ & $0.892$ & $0.910$ & $0.895$\\
Norm.\ conformal & $0.897$ & $0.901$ & $0.895$ & $0.891$ & $0.924$ & $0.891$ & $0.899$ & $0.895$ & $0.898$ & $0.898$ & $0.892$\\
GBM-CQR & $0.896$ & $0.907$ & $0.892$ & $0.900$ & $0.900$ & $0.906$ & $0.906$ & $0.906$ & $0.897$ & $0.893$ & $0.904$\\
MFVI + Conf & $0.891$ & $0.902$ & $0.898$ & $0.886$ & $0.904$ & $0.894$ & $0.899$ & $0.902$ & $0.896$ & $0.899$ & $0.894$\\
Laplace + Conf & $0.895$ & $0.904$ & $0.894$ & $0.902$ & $0.911$ & $0.890$ & $0.898$ & $0.899$ & $0.892$ & $0.900$ & $0.891$\\
Traj-Gauss + Conf & $0.894$ & $0.903$ & $0.897$ & $0.905$ & $0.920$ & $0.890$ & $0.897$ & $0.898$ & $0.896$ & $0.902$ & $0.892$\\
Traj-PCA + Conf & $0.895$ & $0.902$ & $0.901$ & $0.903$ & $0.910$ & $0.892$ & $0.897$ & $0.888$ & $0.893$ & $0.902$ & $0.894$\\
Deep ens. + Conf & $0.892$ & $0.907$ & $0.899$ & $0.898$ & $0.903$ & $0.890$ & $0.897$ & $0.893$ & $0.891$ & $0.901$ & $0.895$\\
MALA + Conf & $0.895$ & $0.899$ & $0.886$ & $0.898$ & $0.899$ & $0.893$ & $0.899$ & $0.900$ & $0.890$ & $0.902$ & $0.892$\\
HMC + Conf & $0.894$ & $0.906$ & $0.890$ & $0.895$ & $0.898$ & $0.893$ & $0.897$ & $0.896$ & $0.893$ & $0.897$ & $0.893$\\
\rowcolor{scoperow}\textbf{SCOPE} & $0.902$ & $0.904$ & $0.904$ & $0.918$ & $0.907$ & $0.898$ & $0.897$ & $0.906$ & $0.891$ & $0.901$ & $0.902$\\
\bottomrule
\end{tabular}}
\end{table}

\begin{table}[htbp]\centering
\caption{\textbf{Nominal-90\% region size: two simulations and nine UCI datasets (coverage in parentheses).} Baselines calibrate their scores on $\Dconf$; SCOPE reuses $\Dguide$. Bold denotes the smallest mean size, including HMC$+$Conf; size should be interpreted together with coverage.}
\label{tab:app-confsize}
\small\setlength{\tabcolsep}{3pt}\renewcommand{\arraystretch}{1.05}
\resizebox{\linewidth}{!}{%
\begin{tabular}{lrrrrrrrrrrr}\toprule
Method & Sim.~1 & Sim.~2 & Concrete & Energy & Yacht & Wine & Abalone & Supercond. & Airfoil & Naval & Protein\\
\midrule
CQR & $1.59\,(0.900)$ & $4.96\,(0.904)$ & $18.42\,(0.893)$ & $1.75\,(0.895)$ & $3.59\,(0.913)$ & $2.25\,(0.897)$ & $6.61\,(0.898)$ & $43.33\,(0.900)$ & $6.07\,(0.904)$ & $0.0032\,(0.895)$ & $14.85\,(0.894)$\\
Abs.-residual conformal & $1.67\,(0.899)$ & $7.72\,(0.904)$ & $21.31\,(0.910)$ & $1.97\,(0.903)$ & $5.75\,(0.918)$ & $2.34\,(0.895)$ & $7.42\,(0.900)$ & $57.09\,(0.903)$ & $6.28\,(0.892)$ & $0.0024\,(0.910)$ & $17.66\,(0.895)$\\
Norm.\ conformal & $1.31\,(0.897)$ & $6.87\,(0.901)$ & $20.22\,(0.895)$ & $5.98\,(0.891)$ & $11.32\,(0.924)$ & $2.21\,(0.891)$ & $6.03\,(0.899)$ & $51.39\,(0.895)$ & $9.45\,(0.898)$ & $0.0108\,(0.898)$ & $15.17\,(0.892)$\\
GBM-CQR & $1.26\,(0.896)$ & $4.88\,(0.907)$ & $26.09\,(0.892)$ & $4.14\,(0.900)$ & $7.49\,(0.900)$ & $\mathbf{2.06\,(0.906)}$ & $6.85\,(0.906)$ & $49.17\,(0.906)$ & $11.62\,(0.897)$ & $0.0412\,(0.893)$ & $16.15\,(0.904)$\\
MFVI + Conf & $\mathbf{1.17\,(0.891)}$ & $3.21\,(0.902)$ & $18.11\,(0.898)$ & $2.18\,(0.886)$ & $11.02\,(0.904)$ & $2.19\,(0.894)$ & $6.00\,(0.899)$ & $43.99\,(0.902)$ & $7.03\,(0.896)$ & $0.0061\,(0.899)$ & $15.28\,(0.894)$\\
Laplace + Conf & $1.26\,(0.895)$ & $3.31\,(0.904)$ & $19.73\,(0.894)$ & $1.77\,(0.902)$ & $3.65\,(0.911)$ & $2.19\,(0.890)$ & $5.98\,(0.898)$ & $40.06\,(0.899)$ & $6.44\,(0.892)$ & $0.0015\,(0.900)$ & $14.99\,(0.891)$\\
Traj-Gauss + Conf & $1.29\,(0.894)$ & $3.47\,(0.903)$ & $19.00\,(0.897)$ & $1.77\,(0.905)$ & $4.07\,(0.920)$ & $2.19\,(0.890)$ & $5.97\,(0.897)$ & $40.78\,(0.898)$ & $6.77\,(0.896)$ & $0.0017\,(0.902)$ & $15.00\,(0.892)$\\
Traj-PCA + Conf & $1.33\,(0.895)$ & $3.69\,(0.902)$ & $18.62\,(0.901)$ & $1.75\,(0.903)$ & $3.60\,(0.910)$ & $2.20\,(0.892)$ & $5.96\,(0.897)$ & $57.07\,(0.888)$ & $6.62\,(0.893)$ & $0.0014\,(0.902)$ & $\mathbf{14.79\,(0.894)}$\\
Deep ens. + Conf & $1.25\,(0.892)$ & $3.03\,(0.907)$ & $16.05\,(0.899)$ & $\mathbf{1.57\,(0.898)}$ & $\mathbf{3.37\,(0.903)}$ & $2.14\,(0.890)$ & $5.90\,(0.897)$ & $35.56\,(0.893)$ & $6.06\,(0.891)$ & $0.0013\,(0.901)$ & $14.80\,(0.895)$\\
MALA + Conf & $1.18\,(0.895)$ & $3.23\,(0.899)$ & $16.96\,(0.886)$ & $1.74\,(0.898)$ & $4.49\,(0.899)$ & $2.17\,(0.893)$ & $5.92\,(0.899)$ & $37.40\,(0.900)$ & $6.37\,(0.890)$ & $0.0014\,(0.902)$ & $14.92\,(0.892)$\\
HMC + Conf & $1.17\,(0.894)$ & $2.88\,(0.906)$ & $\mathbf{15.90\,(0.890)}$ & $1.63\,(0.895)$ & $4.42\,(0.898)$ & $2.15\,(0.893)$ & $\mathbf{5.90\,(0.897)}$ & $\mathbf{32.70\,(0.896)}$ & $\mathbf{5.33\,(0.893)}$ & $\mathbf{0.0011\,(0.897)}$ & $14.91\,(0.893)$\\
\rowcolor{scoperow}\textbf{SCOPE} & $1.21\,(0.902)$ & $\mathbf{2.86\,(0.904)}$ & $18.83\,(0.904)$ & $1.90\,(0.918)$ & $4.44\,(0.907)$ & $2.23\,(0.898)$ & $5.93\,(0.897)$ & $42.10\,(0.906)$ & $6.29\,(0.891)$ & $0.0015\,(0.901)$ & $15.36\,(0.902)$\\
\bottomrule
\end{tabular}}
\end{table}

\begin{table}[htbp]
\centering
\caption{\textbf{Finite-sample prediction-set comparison on the eight main
benchmarks.} Every method uses an independent calibration split and therefore
has the standard finite-sample marginal guarantee under exchangeability.
Coverage reports the maximum absolute deviation from 0.90; size is relative to
HMC+Conf (median [range]); the last column is mean symmetric-difference length
from the population global-level oracle region on Simulation~2. GBM-CQR is gradient-boosted CQR.}
\label{tab:app-finite-sample-summary}
\small\setlength{\tabcolsep}{4pt}
\begin{tabular}{@{}lccc@{}}
\toprule
Method & Coverage max $|$dev.$|$ & Size / HMC+Conf & Sim.~2 $\triangle$ oracle\\
\midrule
CQR & 0.013 & 1.14 [0.81, 1.72] & 3.01\\
Abs.-residual conformal & 0.018 & 1.32 [1.09, 2.68] & 5.92\\
Norm.-residual conformal & 0.024 & 1.42 [1.02, 3.68] & 5.06\\
GBM-CQR & \textbf{0.008} & 1.57 [0.96, 2.55] & 2.89\\
MALA+Conf & 0.014 & 1.04 [1.00, 1.14] & 1.23\\
\rowcolor{scoperow}\textbf{SCOPE/Split} & 0.011 & \textbf{1.02} [1.00, 1.28] & \textbf{0.83}\\
\midrule
Deep ens.+Conf & 0.010 & 1.00 [0.76, 1.09] & 1.01\\
HMC+Conf & 0.010 & 1.00 & 0.81\\
\bottomrule
\end{tabular}
\end{table}

\subsection{Anchor Selection and Baseline Sensitivity}
\label{app:anchor-study}
\label{app:oracle-anchor}

The guide-set NLL chooses a consistent anchor on each benchmark. It selects
the converged MAP on Energy and Yacht in $\numAnchorMapFracEnergy\%$ and
$\numAnchorMapFracYacht\%$ of seeds, and the VI mean on Superconductivity,
Concrete and Wine in $\numAnchorViFracSupercond\%$,
$\numAnchorViFracConcrete\%$ and $\numAnchorViFracWine\%$ of seeds. The choice matters: guide-based selection avoids committing to a single
training-fitted reference across datasets. Table~\ref{tab:app-anchor}
reports the two candidate fits and the selected predictive.

Table~\ref{tab:app-anchor} evaluates the anchor rule
\eqref{eq:anchor-selection}. Its selected test NLLs match the primary
SCOPE predictive in Table~\ref{tab:app-nll}; the candidate and envelope
columns assess the two fixed-anchor strategies.
The candidate references have different predictive behaviour on the
low-noise tasks. Anchor test
RMSE changes from $\numAnchorRmseShortEnergy$ to $\numAnchorRmseConvEnergy$
on Energy and from $\numAnchorRmseShortYacht$ to $\numAnchorRmseConvYacht$
on Yacht after MAP convergence. Before selection, region sizes are
$\numSizeBeforeEnergy\times$, $\numSizeBeforeYacht\times$ and
$\numSizeBeforeNaval\times$ HMC$+$Conf on Energy, Yacht and Naval.
The converged MAP is not uniformly preferable: on Superconductivity its
guide NLL is $\numGuideNllMapSupercond$, versus $\numGuideNllViSupercond$
at the VI mean, and on Concrete it is worse on both guide and test NLL.

Across all nine UCI datasets (six main plus Airfoil, Naval and Protein), guide
selection has mean test NLL no higher than the better fixed-anchor
strategy. The envelope compares dataset-level means and is not used for
selection. The recorded baseline-anchor analysis gives a maximum penalty of
$\numBaselinePenaltyMax$ nats (trajectory Gaussian on Yacht); its SCOPE
mean rank on the six main UCI tasks changes from
$\numRecordedAnchorRankSelected$ to $\numRecordedAnchorRankEnvelope$
when the baselines use their better fixed strategy. These ranks describe
that recorded sensitivity analysis. The primary NLL comparisons use
Table~\ref{tab:app-nll}.

\begin{table}[htbp]\centering
\caption{\textbf{Anchor selection on all nine UCI datasets (20 seeds).} \emph{Short} is the shared staged initialisation, including its 80-epoch MAP stage; \emph{converged} continues it with full-batch Adam and L-BFGS on $\Dtrain$ only. Guide NLL evaluates each candidate's untempered completed predictive on $\Dguide$; \emph{MAP chosen} is the fraction of seeds selecting the converged-MAP candidate. The selected test NLL is the primary SCOPE result in Table~\ref{tab:app-nll}. The envelope is the better of the two fixed-anchor strategies in mean test NLL, a retrospective comparison, and penalty is selected minus envelope. The last two columns give guide-reused region size relative to HMC$+$Conf before and after selection.}
\label{tab:app-anchor}
\small\setlength{\tabcolsep}{3pt}\renewcommand{\arraystretch}{1.05}
\resizebox{\linewidth}{!}{%
\begin{tabular}{l rrr rrr rrrrr rr}\toprule
 & \multicolumn{3}{c}{Anchor} & \multicolumn{3}{c}{Guide NLL} & \multicolumn{5}{c}{Test NLL} & \multicolumn{2}{c}{Size / HMC$+$Conf}\\\cmidrule(lr){2-4}\cmidrule(lr){5-7}\cmidrule(lr){8-12}\cmidrule(lr){13-14}Benchmark & RMSE short & RMSE conv.\ & conv.\ time (s) & VI & MAP & MAP chosen & VI & MAP & selected & envelope & penalty & before & after\\
\midrule
Concrete & 6.647 & 5.806 & 6.0 & 3.145 & 3.467 & 5\% & 3.138 & 3.327 & 3.137 & 3.138 & -0.002 & 1.23 & 1.18\\
Energy & 2.057 & 0.539 & 5.5 & 1.185 & 0.978 & 100\% & 1.204 & 0.810 & 0.810 & 0.810 & 0.000 & 1.89 & 1.17\\
Yacht & 3.332 & 1.176 & 5.2 & 2.261 & 1.606 & 100\% & 2.253 & 1.482 & 1.482 & 1.482 & 0.000 & 2.45 & 1.00\\
Wine & 0.700 & 0.701 & 6.2 & 1.046 & 1.065 & 5\% & 1.051 & 1.077 & 1.051 & 1.051 & 0.000 & 1.04 & 1.04\\
Abalone & 2.112 & 2.108 & 7.7 & 2.009 & 2.002 & 70\% & 2.040 & 2.038 & 2.037 & 2.038 & -0.001 & 1.02 & 1.01\\
Supercond. & 15.252 & 15.896 & 6.7 & 3.849 & 6.192 & 0\% & 3.819 & 7.262 & 3.819 & 3.819 & 0.000 & 1.39 & 1.29\\
Airfoil & 3.024 & 2.198 & 6.4 & 2.147 & 2.089 & 85\% & 2.165 & 2.113 & 1.909 & 2.113 & -0.204 & 1.49 & 1.18\\
Naval & 0.004 & 0.000 & 6.7 & -4.987 & -5.953 & 100\% & -4.985 & -6.954 & -6.954 & -6.954 & 0.000 & 8.11 & 1.37\\
Protein & 4.970 & 4.895 & 5.7 & 3.020 & 3.002 & 70\% & 3.006 & 2.981 & 2.969 & 2.981 & -0.012 & 1.05 & 1.03\\
\bottomrule
\end{tabular}}
\end{table}

\subsection{Rank-Limited Score Guidance on the Main Simulations}
\label{app:sim-guidance-r4}
\label{app:sim-guidance-rank}

\paragraph{Question and design.}
We use the two main simulations to test whether boundary-weighted score
gradients identify sampling directions that improve the \emph{final calibrated
predictive region} when the active rank is small. The primary comparison is
between boundary-weighted and uniformly weighted score operators. Both use the
same score Jacobian and curvature metric; they differ only in how guide
observations are weighted.

For each simulation and seed, guided and uniform constructions share the
training data, guide observations, training-fitted anchor, curvature metric,
main-chain proposal budget, sampler adaptation, projected VI completion and
energy-guard rule. We evaluate the VI and MAP anchors separately and average
their region errors within each seed before constructing paired method
contrasts. Thus the anchor is held fixed within each comparison. Each fitted
predictive receives its own threshold from the same independent calibration
observations, so under exchangeability both constructions have the same
split-conformal marginal coverage guarantee.

\paragraph{Population oracle and outcome.}
Let $p^\star(y\mid x)$ denote the known true conditional density. Define the
population global-level oracle region
\[
C^\star(x)=\{y:p^\star(y\mid x)\ge \tau^\star\},
\]
where the single density level $\tau^\star$ is chosen so that
$\Pr\{Y\in C^\star(X)\}=0.90$. The expectation over the true covariate
distribution used to determine $\tau^\star$ is approximated with
$250{,}000$ draws of $X$. This is the population analogue of SCOPE's use of one
density level across inputs; it is not the input-wise conditional 90\%
highest-density region.

Let $C_{m,s}(x)$ be the independently calibrated region for construction $m$ in
seed $s$. We report normalized symmetric-difference error
\begin{equation}
E_{m,s}=100\,
\frac{\sum_{j=1}^{n_{\mathrm{test}}}
\operatorname{Leb}\!\left(C_{m,s}(x_j)\triangle C^\star(x_j)\right)}
{\sum_{j=1}^{n_{\mathrm{test}}}\operatorname{Leb}\!\left(C^\star(x_j)\right)}.
\label{eq:sim-guidance-error}
\end{equation}
The metric is the percentage of the oracle regions' total length on which the
calibrated fitted region disagrees with the population oracle. It detects
misplaced boundaries, omitted components and unnecessary inclusion of
low-density gaps.

\paragraph{Rank-four comparison.}
Table~\ref{tab:bwc-main} reports the rank-$\numSimGuidanceRank$ comparison over
$\numSeedCount$ matched seeds. Boundary weighting reduces mean normalized
oracle-region error from $\numSimOneUniformErr\%$ to
$\numSimOneGuidedErr\%$ on Simulation~1 and from
$\numSimTwoUniformErrRfour\%$ to $\numSimTwoGuidedErrRfour\%$ on
Simulation~2, approximately $22\%$ in each case. The guided-minus-uniform
paired mean differences are $\numSimOneDiff$ percentage points with
multiplicity-adjusted 95\% interval
$[\numSimOneDiffLo,\numSimOneDiffHi]$ and $\numSimTwoDiff$ percentage points
with interval $[\numSimTwoDiffLo,\numSimTwoDiffHi]$, respectively.

\paragraph{Rank sweep.}
We repeat the matched guided-versus-uniform comparison at
$r\in\{2,4,8,16,32\}$. Table~\ref{tab:guidance-rank-sweep} reports the
absolute mean errors and paired uncertainty. In both simulations, the reduction
in mean error is approximately $\numGuidanceSweepRTwo\%$,
$\numGuidanceSweepRFour\%$, $\numGuidanceSweepREight\%$,
$\numGuidanceSweepRSixteen\%$ and $\numGuidanceSweepRThirtyTwo\%$ at ranks
$2,4,8,16$ and $32$, respectively. The adjusted intervals exclude zero through
rank $16$ on both simulations and include zero at rank $32$ on both; the latter
shows that the guided advantage is statistically unresolved there, not that the
two constructions are equivalent.

\begin{table}[!htbp]
\centering
\caption{\textbf{Score-guidance benefit across active rank.}
Normalized symmetric-difference error (\%) relative to the population
global-level oracle region for guided and uniform weighting under the
fixed-anchor protocol. Within each seed, the VI- and MAP-anchor errors are
averaged before comparison. Differences are guided minus uniform, with
multiplicity-adjusted paired 95\% bootstrap intervals over 20 matched seeds.
Lower is better.}
\label{tab:guidance-rank-sweep}
\tiny
\setlength{\tabcolsep}{3.2pt}
\resizebox{\columnwidth}{!}{%
\begin{tabular}{@{}lrrrr@{}}
\toprule
Simulation & Rank & Uniform & Guided &
Guided $-$ uniform [95\% CI] \\
\midrule
Simulation~1
 & 2  & 21.4 & 15.0 & $-6.4\;[-9.2,-3.5]$ \\
 & 4  & 16.7 & 13.0 & $-3.7\;[-5.6,-1.8]$ \\
 & 8  & 13.7 & 11.5 & $-2.2\;[-3.4,-1.0]$ \\
 & 16 & 11.3 & 10.3 & $-1.0\;[-1.8,-0.2]$ \\
 & 32 & 10.2 &  9.9 & $-0.3\;[-0.9,+0.3]$ \\
\midrule
Simulation~2
 & 2  & 57.1 & 40.0 & $-17.1\;[-24.0,-10.2]$ \\
 & 4  & 47.4 & 37.0 & $-10.4\;[-15.6,-5.1]$ \\
 & 8  & 42.3 & 35.5 & $-6.8\;[-10.5,-3.0]$ \\
 & 16 & 37.7 & 34.3 & $-3.4\;[-6.0,-0.8]$ \\
 & 32 & 34.5 & 33.5 & $-1.0\;[-3.0,+1.0]$ \\
\bottomrule
\end{tabular}}
\end{table}

The fixed-anchor protocol is designed to isolate direction selection; these
results do not by themselves establish the same advantage after method-specific
anchor selection. The default spectral-mass rule selects rank adaptively rather than fixing it;
the rank sweep is therefore a controlled study of direction-selection quality
at matched rank. The comparisons match the main-chain proposal budget rather than total
wall-clock time and therefore isolate how efficiently a given active dimension
is allocated.

\subsection{Completion and Guard}
\label{app:completion}

H2 compares active-only, unguarded and guarded completion on the same
retained active states. The completed NLLs match the primary SCOPE
predictive in Table~\ref{tab:app-nll}.
\ifCompletionReady
Table~\ref{tab:app-completion} covers $\numHTwoN$ of the eight main benchmarks: Simulation~1 and the six main UCI datasets (Simulation~2 excluded). Guide-reused coverage differences are at most $\numHTwoCovDiffMaxAbs$, while completion lowers mean NLL on all
$\numHTwoNActiveWorse$ benchmarks by up to $\numHTwoNllGainMax$ nats.
On Energy and Yacht, these predictive improvements accompany decreases
in TV to HMC from
$\numHTwoTvEnergyActive$ to $\numHTwoTvEnergyCompleted$ and from
$\numHTwoTvYachtActive$ to $\numHTwoTvYachtCompleted$, respectively.

Without the guard, the VI-scale complement increases region size by up to
$\numHTwoUnguardSizeRatioMax\times$ ($\numHTwoUnguardSizeRatioMaxName$) and
NLL by up to $\numHTwoUnguardNllPenaltyMax$ nats. Naval also exhibits
occasional non-finite unguarded proposals. Table~\ref{tab:app-diag} gives
the guard scales and energy diagnostics. These results assess the guard
empirically; Proposition~\ref{prop:gaussian} applies before this modification.
\begin{table}[htbp]\centering
\caption{\textbf{Completion ablation (H2): Simulation~1 and six main UCI datasets; matched active states and seeds.} Active-only uses the retained active states without a complement ($K=1$); unguarded adds the projected VI complement without the energy guard; completed is the guarded SCOPE predictive. The completed NLLs match the primary SCOPE results in Table~\ref{tab:app-nll}. Coverage and size use the guide-reused region. NLL, TV (to the oracle or HMC reference), and model-region coverage use the untempered predictive.}
\label{tab:app-completion}
\small\setlength{\tabcolsep}{2.6pt}\renewcommand{\arraystretch}{1.05}
\resizebox{\linewidth}{!}{%
\begin{tabular}{l rrr rrr rrr rrr rrr}\toprule
 & \multicolumn{3}{c}{Coverage} & \multicolumn{3}{c}{Size} & \multicolumn{3}{c}{NLL} & \multicolumn{3}{c}{TV} & \multicolumn{3}{c}{Model-region cov.}\\ \cmidrule(lr){2-4}\cmidrule(lr){5-7}\cmidrule(lr){8-10}\cmidrule(lr){11-13}\cmidrule(lr){14-16}Benchmark & active & unguarded & completed & active & unguarded & completed & active & unguarded & completed & active & unguarded & completed & active & unguarded & completed\\
\midrule
Sim.~1 & 0.900 & 0.902 & 0.902 & 1.20 & 1.20 & 1.21 & 0.379 & 0.378 & 0.368 & 0.069 & 0.070 & 0.070 & 0.903 & 0.913 & 0.913\\
Concrete & 0.905 & 0.901 & 0.904 & 18.67 & 18.89 & 18.83 & 3.138 & 3.154 & 3.137 & 0.178 & 0.205 & 0.193 & 0.892 & 0.925 & 0.918\\
Energy & 0.917 & 0.902 & 0.918 & 1.84 & 4.26 & 1.90 & 0.836 & 1.984 & 0.810 & 0.132 & 0.537 & 0.109 & 0.956 & 1.000 & 0.983\\
Yacht & 0.913 & 0.903 & 0.907 & 3.91 & 13.84 & 4.44 & 1.523 & 2.831 & 1.482 & 0.213 & 0.524 & 0.109 & 0.941 & 1.000 & 0.984\\
Wine & 0.897 & 0.899 & 0.898 & 2.22 & 2.23 & 2.23 & 1.057 & 1.053 & 1.051 & 0.060 & 0.064 & 0.061 & 0.898 & 0.909 & 0.904\\
Abalone & 0.897 & 0.900 & 0.897 & 5.92 & 6.30 & 5.93 & 2.045 & 2.139 & 2.037 & 0.048 & 0.161 & 0.046 & 0.899 & 0.949 & 0.912\\
Supercond. & 0.902 & 0.906 & 0.906 & 40.54 & 42.10 & 42.10 & 4.013 & 3.819 & 3.819 & 0.234 & 0.263 & 0.263 & 0.940 & 0.966 & 0.966\\
\bottomrule
\end{tabular}}
\end{table}

\else
Same-chain completion comparisons are unavailable in this data state.
\fi

\subsection{Independent Calibration and Guide Size}
\label{app:split-fix}
\label{app:guide-size}

SCOPE/Split freezes the predictive and calibrates its density score on
independent $\Dconf$.
\ifSplitReady
Table~\ref{tab:app-split} reports realised coverage
$\numSplitCovLo$--$\numSplitCovHi$, a paired coverage difference from
guide-reused SCOPE of at most $\numReuseGapMaxAbs$, and size ratio
$\numReuseSizeRatioLo$--$\numReuseSizeRatioHi$. The tempered split variant
also calibrates and evaluates the same frozen predictive.
\begin{table}[htbp]\centering
\caption{\textbf{Guide reuse versus an independent calibration split (20 seeds).} SCOPE (level from $\Dguide$) and SCOPE/Split (level from $\Dconf$) on the same untempered predictive; `gap' is the empirical paired coverage difference SCOPE $-$ SCOPE/Split; it is not an estimate of the theoretical coverage penalty in
Theorem~\ref{thm:reuse}. SCOPE/Split (tempered) calibrates and evaluates the same guide-tempered predictive. HMC$+$Conf is shown for reference.}
\label{tab:app-split}
\small\setlength{\tabcolsep}{3pt}\renewcommand{\arraystretch}{1.05}
\resizebox{\linewidth}{!}{%
\begin{tabular}{l rrrrr rrrr}\toprule
 & \multicolumn{5}{c}{Coverage} & \multicolumn{4}{c}{Size}\\ \cmidrule(lr){2-6}\cmidrule(lr){7-10}Benchmark & SCOPE & SCOPE/Split & gap & Split (tempered) & HMC$+$Conf & SCOPE & SCOPE/Split & ratio & HMC$+$Conf\\
\midrule
Sim.~1 & 0.902 & 0.893 & +0.009 & 0.893 & 0.894 & 1.21 & 1.17 & 1.026 & 1.17\\
Sim.~2 & 0.904 & 0.905 & -0.001 & 0.909 & 0.906 & 2.86 & 2.89 & 0.989 & 2.88\\
Concrete & 0.904 & 0.898 & +0.006 & 0.899 & 0.890 & 18.83 & 18.33 & 1.027 & 15.90\\
Energy & 0.918 & 0.895 & +0.023 & 0.898 & 0.895 & 1.90 & 1.80 & 1.054 & 1.63\\
Yacht & 0.907 & 0.911 & -0.004 & 0.913 & 0.898 & 4.44 & 4.42 & 1.004 & 4.42\\
Wine & 0.898 & 0.894 & +0.004 & 0.894 & 0.893 & 2.23 & 2.20 & 1.016 & 2.15\\
Abalone & 0.897 & 0.899 & -0.002 & 0.898 & 0.897 & 5.93 & 5.96 & 0.995 & 5.90\\
Supercond. & 0.906 & 0.904 & +0.002 & 0.902 & 0.896 & 42.10 & 41.98 & 1.003 & 32.70\\
\bottomrule
\end{tabular}}
\end{table}

\fi

\ifGuideSizeReady
Table~\ref{tab:app-guidesize-full} reruns the complete guide-dependent
pipeline at each guide size on a $\numHThreeNds$-dataset subset of the main UCI suite: Concrete, Energy, Yacht and Superconductivity. The paired coverage difference from SCOPE/Split ranges from $\numHThreeGapMin$ to $+\numHThreeGapMax$ and is at most $\numHThreeGapMaxAtLeastHundred$ for
$m_g\ge100$. The largest conservative difference occurs at
$m_g=\numHThreeSmallest$. These are empirical coverage comparisons, not numerical estimates of
the complexity penalty in Theorem~\ref{thm:reuse}.
\begin{table}[htbp]\centering
\caption{\textbf{Guide-size study (H3): the whole SCOPE pipeline rerun on random subsamples of $\Dguide$ of size $m_g$ (20 seeds).} For each size: coverage of the guide-reused region (SCOPE), coverage of the split-conformal region on the untouched $\Dconf$ (Split) from the same run, the size ratio SCOPE/Split, and the mean squared projection of the active basis at $m_g$ onto the basis obtained with the full guide set at the same anchor (overlap). Anchor selection, geometry, pilot refinement and level all use only the $m_g$ points.}
\label{tab:app-guidesize-full}
\small\setlength{\tabcolsep}{3pt}\renewcommand{\arraystretch}{1.05}
\resizebox{\linewidth}{!}{%
\begin{tabular}{l rrrr rrrr rrrr rrrr}\toprule
 & \multicolumn{4}{c}{Concrete ($m_g^{\mathrm{full}}=206$)} & \multicolumn{4}{c}{Energy ($m_g^{\mathrm{full}}=154$)} & \multicolumn{4}{c}{Supercond. ($m_g^{\mathrm{full}}=400$)} & \multicolumn{4}{c}{Yacht ($m_g^{\mathrm{full}}=62$)}\\\cmidrule(lr){2-5}\cmidrule(lr){6-9}\cmidrule(lr){10-13}\cmidrule(lr){14-17}$m_g$ & SCOPE & Split & size & overlap & SCOPE & Split & size & overlap & SCOPE & Split & size & overlap & SCOPE & Split & size & overlap\\
\midrule
25 & 0.912 & 0.893 & 1.13 & 0.10 & 0.918 & 0.894 & 1.14 & 0.14 & 0.931 & 0.902 & 1.17 & 0.08 & 0.917 & 0.911 & 1.07 & 0.29\\
50 & 0.909 & 0.895 & 1.06 & 0.25 & 0.909 & 0.892 & 1.05 & 0.25 & 0.908 & 0.905 & 1.06 & 0.10 & 0.907 & 0.912 & 1.00 & 0.60\\
62 & -- & -- & -- & -- & -- & -- & -- & -- & -- & -- & -- & -- & 0.907 & 0.911 & 1.00 & --\\
100 & 0.896 & 0.900 & 1.01 & 0.47 & 0.904 & 0.893 & 1.03 & 0.50 & 0.904 & 0.904 & 1.00 & 0.21 & -- & -- & -- & --\\
154 & -- & -- & -- & -- & 0.918 & 0.895 & 1.05 & -- & -- & -- & -- & -- & -- & -- & -- & --\\
200 & 0.904 & 0.896 & 1.03 & 0.94 & -- & -- & -- & -- & 0.910 & 0.904 & 1.02 & 0.44 & -- & -- & -- & --\\
206 & 0.904 & 0.898 & 1.03 & -- & -- & -- & -- & -- & -- & -- & -- & -- & -- & -- & -- & --\\
400 & -- & -- & -- & -- & -- & -- & -- & -- & 0.906 & 0.904 & 1.00 & -- & -- & -- & -- & --\\
\bottomrule
\end{tabular}}
\end{table}

\else
The full-pipeline guide-size study is unavailable in this data state.
\fi
The result files also contain a level-only study that freezes the score
rule and subsamples the guide observations. It measures quantile sensitivity, not the effect of refitting the
complete guide-dependent procedure.

\subsection{Efficiency and Paired Comparisons}
\label{app:functionals}
\label{app:ess-percentiles}
\label{app:paired}

The coordinate diagnostics use a score-derived task basis. We additionally
report the ESS of the fixed-level smooth-membership functional and log
predictive densities at $32$ fixed test probes, using sequences of
$2{,}000$ retained states and the initial-positive-sequence estimator.
Antithetic components are averaged within each state.
\ifFunctionalsReady
Boundary-functional and probe ESS/s are
$\numBoundaryEssVsMalaLo$--$\numBoundaryEssVsMalaHi\times$ and
$\numProbeEssVsMalaLo$--$\numProbeEssVsMalaHi\times$ MALA's across all nine
UCI datasets\ifHmcRerunReady, and $\numBoundaryEssVsHmcLo$--$\numBoundaryEssVsHmcHi\times$ and
$\numProbeEssVsHmcLo$--$\numProbeEssVsHmcHi\times$ the rerun HMC reference's
on the six main UCI datasets\fi.
Against matched PCA, the corresponding ratios are
$\numHOneBoundaryVsPcaLo$--$\numHOneBoundaryVsPcaHi\times$ and
$\numHOneProbeVsPcaLo$--$\numHOneProbeVsPcaHi\times$ on the six main UCI datasets.
Thus the advantage is smaller for these basis-free functionals than for
the coordinate summaries.
\begin{table}[htbp]\centering
\caption{\textbf{Chain efficiency for common predictive functionals (per fit second; means over seeds).} Boundary: ESS of the common boundary functional defined in Appendix~\ref{app:metrics} (mean smoothed score membership on guide observations at a fixed reference threshold). Probe: 90th percentile over 32 fixed test probes of the ESS of $\log p_s(y^\star_j\mid x^\star_j)$ along retained draws. SCOPE is the guide-selected run; PCA the matched control at the converged-MAP anchor; MALA is rerun at its guide-selected anchor with the published seeds; HMC is the reference chain rerun on $n_{\mathrm{HMC}}$ seeds (matched seeds and sampler configuration). MALA-relative ratios cover all nine UCI datasets; HMC-relative ratios cover only the six main UCI datasets. All estimates use the initial-positive-sequence estimator.}
\label{tab:app-functionals}
\small\setlength{\tabcolsep}{2.8pt}\renewcommand{\arraystretch}{1.05}
\resizebox{\linewidth}{!}{%
\begin{tabular}{l rrrrrr rrrrrr r}\toprule
 & \multicolumn{6}{c}{Boundary-functional ESS/s} & \multicolumn{6}{c}{Predictive-probe ESS/s} & \\ \cmidrule(lr){2-7}\cmidrule(lr){8-13}Benchmark & SCOPE & PCA & MALA & HMC & vs MALA & vs HMC & SCOPE & PCA & MALA & HMC & vs MALA & vs HMC & $n_{\mathrm{HMC}}$\\
\midrule
Sim.~1 & 19.70 & 9.21 & -- & -- & -- & -- & 23.11 & 17.95 & -- & -- & -- & -- & --\\
Sim.~2 & 4.72 & -- & -- & -- & -- & -- & -- & -- & -- & -- & -- & -- & --\\
Concrete & 34.23 & 25.28 & 0.35 & 0.25 & 96$\times$ & 135$\times$ & 43.80 & 44.14 & 0.77 & 0.62 & 57$\times$ & 71$\times$ & 20\\
Energy & 36.33 & 34.51 & 0.36 & 0.56 & 101$\times$ & 65$\times$ & 39.27 & 40.93 & 1.06 & 0.93 & 37$\times$ & 42$\times$ & 20\\
Yacht & 38.84 & 31.78 & 0.43 & 1.16 & 89$\times$ & 33$\times$ & 40.35 & 42.15 & 1.11 & 1.51 & 36$\times$ & 27$\times$ & 20\\
Wine & 15.82 & 7.94 & 0.53 & 0.33 & 30$\times$ & 48$\times$ & 22.44 & 12.95 & 0.74 & 0.58 & 30$\times$ & 39$\times$ & 20\\
Abalone & 23.31 & 8.15 & 0.44 & 0.72 & 53$\times$ & 32$\times$ & 28.37 & 21.32 & 0.74 & 1.01 & 38$\times$ & 28$\times$ & 20\\
Supercond. & 28.17 & 29.89 & 0.24 & 0.09 & 118$\times$ & 309$\times$ & 33.79 & 37.13 & 0.45 & 0.39 & 76$\times$ & 87$\times$ & 20\\
Airfoil & 33.26 & 8.83 & 0.30 & -- & 111$\times$ & -- & 41.25 & 29.67 & 0.64 & -- & 64$\times$ & -- & --\\
Naval & 28.51 & 15.52 & 0.26 & -- & 109$\times$ & -- & 31.34 & 34.78 & 1.23 & -- & 25$\times$ & -- & --\\
Protein & 26.62 & 19.00 & 0.79 & -- & 34$\times$ & -- & 29.10 & 22.62 & 0.93 & -- & 31$\times$ & -- & --\\
\bottomrule
\end{tabular}}
\end{table}

\fi

Table~\ref{tab:app-ess-percentiles} reports the coordinate comparison at
additional percentiles. On the six main UCI datasets, the SCOPE/HMC ratio ranges from
$\numEssRatioHmcPNinetyLo$ to $\numEssRatioHmcPNinetyHi$ at the 90th
percentile, $\numEssRatioHmcMedianLo$ to $\numEssRatioHmcMedianHi$ at the
median, and $\numEssRatioHmcMinLo$ to $\numEssRatioHmcMinHi$ at the minimum;
the smallest reported ratio across these summaries is
$\numEssRatioHmcAnyPercentileMin$.
\begin{table}[htbp]\centering
\caption{\textbf{Sensitivity of coordinate ESS/s to the coordinate summary.} For each seed, ESS is summarised at the stated percentile over the 20 common-basis coordinates; the mean of these counts is divided by mean wall-clock. SCOPE uses the two-candidate inference time of Table~\ref{tab:main}. The table reports these rates and their SCOPE/HMC and SCOPE/MALA ratios. Only UCI benchmarks with comparable timing cohorts are included. The main comparison uses the same 90th coordinate percentile, but averages seedwise ESS/s rates before taking method ratios, so its values need not equal this table's ratios of means (Appendix~\ref{app:metrics}). These summaries assess autocorrelation in the common basis, not accuracy of the posterior approximation.}
\label{tab:app-ess-percentiles}
\scriptsize\setlength{\tabcolsep}{2.2pt}\renewcommand{\arraystretch}{1.05}
\resizebox{\linewidth}{!}{%
\begin{tabular}{l rrrr rrrr rrrr rrrr}\toprule
 & \multicolumn{4}{c}{min over coordinates} & \multicolumn{4}{c}{10th percentile} & \multicolumn{4}{c}{median} & \multicolumn{4}{c}{90th percentile}\\ \cmidrule(lr){2-5}\cmidrule(lr){6-9}\cmidrule(lr){10-13}\cmidrule(lr){14-17}Benchmark & SCOPE & HMC & /HMC & /MALA & SCOPE & HMC & /HMC & /MALA & SCOPE & HMC & /HMC & /MALA & SCOPE & HMC & /HMC & /MALA\\
\midrule
Concrete & 10.2 & 0.010 & 994$\times$ & 142$\times$ & 10.9 & 0.015 & 736$\times$ & 141$\times$ & 13.6 & 0.030 & 453$\times$ & 118$\times$ & 15.8 & 0.057 & 278$\times$ & 73$\times$\\
Energy & 11.7 & 0.009 & 1303$\times$ & 161$\times$ & 12.7 & 0.012 & 1024$\times$ & 158$\times$ & 16.2 & 0.025 & 647$\times$ & 142$\times$ & 20.0 & 0.048 & 412$\times$ & 112$\times$\\
Yacht & 21.4 & 0.358 & 60$\times$ & 154$\times$ & 21.8 & 0.372 & 59$\times$ & 151$\times$ & 23.5 & 0.426 & 55$\times$ & 141$\times$ & 25.5 & 0.488 & 52$\times$ & 118$\times$\\
Wine & 4.2 & 0.015 & 270$\times$ & 75$\times$ & 4.9 & 0.022 & 222$\times$ & 75$\times$ & 6.1 & 0.040 & 150$\times$ & 58$\times$ & 7.3 & 0.065 & 112$\times$ & 40$\times$\\
Abalone & 5.3 & 0.016 & 340$\times$ & 106$\times$ & 6.2 & 0.023 & 272$\times$ & 111$\times$ & 7.9 & 0.050 & 159$\times$ & 91$\times$ & 9.7 & 0.091 & 107$\times$ & 64$\times$\\
Supercond. & 5.2 & 0.011 & 458$\times$ & 84$\times$ & 5.7 & 0.015 & 368$\times$ & 86$\times$ & 7.1 & 0.029 & 242$\times$ & 72$\times$ & 8.4 & 0.057 & 147$\times$ & 49$\times$\\
Airfoil & 8.9 & 0.005 & 1853$\times$ & 132$\times$ & 10.3 & 0.006 & 1656$\times$ & 136$\times$ & 12.8 & 0.011 & 1165$\times$ & 115$\times$ & 15.9 & 0.024 & 649$\times$ & 86$\times$\\
Naval & 5.7 & 0.004 & 1295$\times$ & 83$\times$ & 5.9 & 0.005 & 1219$\times$ & 82$\times$ & 6.8 & 0.006 & 1060$\times$ & 73$\times$ & 7.8 & 0.011 & 684$\times$ & 49$\times$\\
Protein & 10.2 & 0.134 & 76$\times$ & 140$\times$ & 11.2 & 0.190 & 59$\times$ & 126$\times$ & 12.9 & 0.298 & 43$\times$ & 85$\times$ & 14.2 & 0.425 & 33$\times$ & 56$\times$\\
\bottomrule
\end{tabular}}
\end{table}

\begin{table}[htbp]\centering
\caption{\textbf{Predictive ESS for every chain on the eight main benchmarks (20 seeds).} For each seed and method, the initial-positive-sequence ESS of $\log p(y_j\mid x_j,\theta_s)$ is computed along the $2{,}000$ retained draws at every test point $j$ and summarised by its median over the test set in (a) and by its minimum, the slowest-mixing test point, in (b). Effective draws are means over seeds. ESS/s divides the mean count by the method's mean wall-clock, and the ratio is taken to the same quantity for HMC, so (b) reports the relative min-ESS/s. UCI rates charge both SCOPE candidate fits and converged-MAP construction,
and one selected baseline fit, excluding shared initialisation and test
evaluation. Simulation ratios use reconstructed SCOPE times: recorded
single-anchor time multiplied by the measured full-to-single-fit cost ratio
($1.9$ and $2.3$). They are descriptive estimates under differing machine
loads and are excluded from headline efficiency summaries. Each SCOPE draw combines a state of its MCMC chain in the score-guided subspace with an independent draw of the remaining coordinates from the variational completion. \emph{Full}: SCOPE's draws as reported; \emph{MCMC only}: the chain states alone, with the remaining coordinates fixed at the anchor and no completion draws. Laplace, MFVI and the trajectory Gaussian draw independently, so their ESS equals the number of draws by construction. Dashes denote unavailable slowest-point records.}
\label{tab:app-predess}
\small\setlength{\tabcolsep}{4pt}\renewcommand{\arraystretch}{1.05}
\resizebox{\linewidth}{!}{%
\begin{tabular}{l rrrrrr@{\hspace{14pt}}rrrrr}\toprule
 & \multicolumn{6}{c}{Effective draws of $2{,}000$} & \multicolumn{5}{c}{ESS/s relative to HMC}\\\cmidrule(lr){2-7}\cmidrule(lr){8-12}
 & \multicolumn{2}{c}{SCOPE} & & & & & \multicolumn{2}{c}{SCOPE} & & &\\\cmidrule(lr){2-3}\cmidrule(lr){8-9}
Benchmark & Full & MCMC only & HMC & MALA & SGLD & PCA & Full & MCMC only & MALA & SGLD & PCA\\
\midrule
\multicolumn{12}{@{}l}{\emph{(a) Median over test points}}\\[1pt]
Sim.~1 & 1{,}622 & 1{,}323 & 1{,}676 & 32 & 84 & 18 & 16$\times$ & 13$\times$ & 0.43$\times$ & 1.2$\times$ & 0.39$\times$\\
Sim.~2 & 1{,}867 & 1{,}184 & 123 & 76 & 99 & 43 & 47$\times$ & 30$\times$ & 2.3$\times$ & 36$\times$ & 4.7$\times$\\
Concrete & 1{,}757 & 1{,}254 & 382 & 17 & 94 & 22 & 56$\times$ & 40$\times$ & 0.98$\times$ & 5.4$\times$ & 1.8$\times$\\
Energy & 1{,}943 & 1{,}319 & 878 & 26 & 82 & 37 & 33$\times$ & 22$\times$ & 0.73$\times$ & 1.9$\times$ & 1.7$\times$\\
Yacht & 1{,}985 & 1{,}710 & 1{,}642 & 29 & 86 & 28 & 17$\times$ & 15$\times$ & 0.41$\times$ & 1.0$\times$ & 0.64$\times$\\
Wine & 1{,}181 & 731 & 502 & 22 & 78 & 53 & 28$\times$ & 17$\times$ & 0.97$\times$ & 4.4$\times$ & 4.1$\times$\\
Abalone & 1{,}478 & 942 & 1{,}026 & 22 & 82 & 21 & 19$\times$ & 12$\times$ & 0.53$\times$ & 2.6$\times$ & 0.80$\times$\\
Supercond. & 1{,}800 & 829 & 257 & 12 & 74 & 15 & 89$\times$ & 41$\times$ & 1.0$\times$ & 6.5$\times$ & 2.2$\times$\\
\midrule
\multicolumn{12}{@{}l}{\emph{(b) Minimum over test points (slowest-mixing point): relative min-ESS/s}}\\[1pt]
Sim.~1 & 919 & 758 & 590 & 6 & -- & 4 & 25$\times$ & 21$\times$ & 0.22$\times$ & -- & 0.28$\times$\\
Sim.~2 & 1{,}152 & 692 & -- & -- & -- & -- & 89$\times$ & -- & -- & -- & --\\
Concrete & 992 & 781 & 49 & 4 & -- & 6 & 250$\times$ & 196$\times$ & 1.9$\times$ & -- & 4.1$\times$\\
Energy & 1{,}138 & 751 & 212 & 6 & -- & 9 & 79$\times$ & 52$\times$ & 0.67$\times$ & -- & 1.7$\times$\\
Yacht & 1{,}448 & 1{,}303 & 795 & 8 & -- & 9 & 26$\times$ & 23$\times$ & 0.24$\times$ & -- & 0.45$\times$\\
Wine & 487 & 339 & 44 & 4 & -- & 16 & 132$\times$ & 92$\times$ & 2.2$\times$ & -- & 14$\times$\\
Abalone & 669 & 513 & 225 & 5 & -- & 5 & 40$\times$ & 31$\times$ & 0.50$\times$ & -- & 0.85$\times$\\
Supercond. & 819 & 385 & 30 & 4 & -- & 4 & 353$\times$ & 166$\times$ & 2.8$\times$ & -- & 5.3$\times$\\
\bottomrule
\end{tabular}}
\end{table}

\ifPairedReady
Table~\ref{tab:app-paired} gives seed-level paired bootstrap intervals for
region size, coverage and NLL relative to HMC, and the H1 task-mESS/s ratio. The latter uses only finite paired log-ratios;
the available seed counts are shown and can be much smaller than 20.
The result files also retain paired contrasts for guide-versus-split
coverage and completion.
\begin{table}[htbp]\centering
\caption{\textbf{Seed-level paired comparisons: mean paired difference or geometric-mean ratio with 95\% bootstrap confidence intervals.} Size and coverage compare SCOPE with HMC$+$Conf; NLL compares SCOPE with HMC. Differences are SCOPE minus the reference, and size is SCOPE/reference. The first three comparisons use 20 paired seeds. The task-mESS/s ratio compares SCOPE geometry with matched trajectory PCA (H1), using only the $n_{\rm H1}$ finite paired log-ratios shown; non-finite contrasts are excluded. Each contrast pairs methods on the same seed, split and initialisation. The small available sample counts limit the precision of some efficiency comparisons.}
\label{tab:app-paired}
\small\setlength{\tabcolsep}{4pt}\renewcommand{\arraystretch}{1.05}
\resizebox{\linewidth}{!}{%
\begin{tabular}{lccccc}\toprule
Benchmark & Size ratio & $\Delta$coverage & $\Delta$NLL & Task-mESS/s ratio & $n_{\rm H1}$\\
\midrule
Sim.~1 & 1.03 [1.00, 1.05] & +0.008 [-0.001, +0.017] & -0.008 [-0.021, +0.005] & 7.49 [6.19, 8.95] & 18\\
Sim.~2 & 0.99 [0.98, 1.00] & -0.002 [-0.009, +0.005] & -0.034 [-0.051, -0.017] & -- & --\\
Concrete & 1.18 [1.14, 1.23] & +0.014 [+0.003, +0.025] & +0.193 [+0.173, +0.213] & 3.81 [2.87, 4.93] & 12\\
Energy & 1.16 [1.10, 1.22] & +0.023 [+0.006, +0.040] & -0.071 [-0.085, -0.057] & 3.95 [2.77, 5.79] & 18\\
Yacht & 1.00 [0.94, 1.05] & +0.009 [-0.014, +0.029] & -0.211 [-0.236, -0.186] & 3.51 [2.02, 6.29] & 18\\
Wine & 1.04 [1.02, 1.05] & +0.005 [+0.001, +0.010] & +0.023 [+0.019, +0.027] & 5.86 [4.75, 7.24] & 17\\
Abalone & 1.01 [0.98, 1.04] & +0.000 [-0.009, +0.009] & +0.012 [+0.009, +0.015] & 9.68 [8.10, 10.98] & 10\\
Supercond. & 1.29 [1.26, 1.31] & +0.010 [+0.000, +0.019] & +0.266 [+0.240, +0.292] & 2.24 [1.64, 3.05] & 16\\
Airfoil & 1.18 [1.11, 1.24] & -0.002 [-0.015, +0.010] & +0.024 [+0.017, +0.031] & 8.52 [6.54, 11.37] & 8\\
Naval & 1.38 [1.24, 1.52] & +0.004 [-0.010, +0.017] & -0.903 [-0.916, -0.890] & 3.64 [1.57, 8.25] & 7\\
Protein & 1.03 [1.00, 1.06] & +0.009 [-0.003, +0.021] & -0.008 [-0.0096, -0.0064] & 19.21 [15.06, 28.89] & 3\\
\bottomrule
\end{tabular}}
\end{table}

\else
Paired comparisons are unavailable in this data state.
\fi

\subsection{Sensitivity Analyses}
\label{app:alpha}
\label{app:rank}
\label{app:bandwidth}

\begin{figure}[!t]
\centering
\includegraphics[width=\textwidth]
{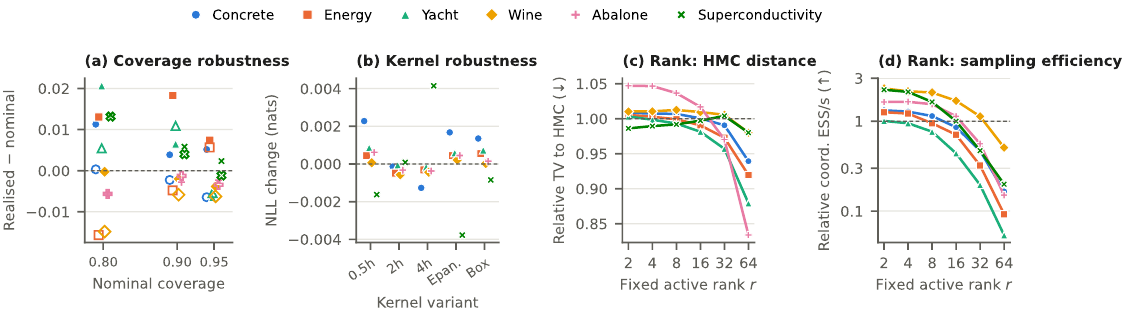}
\caption{\textbf{Sensitivity of \SCOPE\ on the six main UCI
datasets} ($20$ seeds each).
(a)~Realised minus nominal coverage after rerunning the complete procedure at
nominal levels $0.80$, $0.90$, and $0.95$; filled markers denote the
guide-reused \SCOPE\ region and hollow markers denote \SCOPE/Split.
(b)~Test-NLL change under boundary-bandwidth and kernel-shape perturbations.
(c,d)~Sensitivity to fixed active rank, normalised by each dataset's deployed
spectral-mass construction: TV distance to the finite-budget HMC predictive
(c; lower indicates closer agreement with the HMC baseline) and coordinate ESS/s
(d; higher is better; log scale). Dashed lines mark the corresponding default.
The rank panels vary rank within the guided construction and are distinct from
the matched-rank guided-versus-uniform experiment in
Appendix~\ref{app:sim-guidance-rank}.}
\label{fig:app-sensitivity}
\end{figure}

Figure~\ref{fig:app-sensitivity} summarises sensitivity to the nominal
coverage level, boundary-kernel specification, and active rank on the six
main UCI datasets.

\ifAlphaReady
Changing the nominal coverage level rebuilds the kernel, pilot refinement,
basis, and final level. At nominal $0.80$, $0.90$, and $0.95$, realised
coverage remains close to target: the maximum deviation is
$\numAlphaCovMaxDev$ for \SCOPE\ and $\numAlphaSplitCovMaxDev$ for
\SCOPE/Split.
\fi

\ifBandwidthReady
The construction is also insensitive to the precise boundary-kernel
specification. Multiplying the default bandwidth by $0.5$, $2$, or $4$, or
replacing the Gaussian kernel by Epanechnikov or boxcar weights, changes NLL
by at most $\numBwNllMaxDev$ nats and coverage by at most
$\numBwCovMaxDev$.
\fi

\ifRankReady
Fixing the guided active rank at $r\in\{2,4,8,16,32,64\}$ reveals an
efficiency--agreement trade-off. Test NLL changes by at most
$\numRankNllMaxDev$ nats, coverage by at most $\numRankCovMaxDev$, and
region size varies by only a few percent across ranks. At larger ranks, the
predictive generally moves closer in TV distance to the fixed-budget HMC
baseline, while coordinate ESS/s decreases markedly
(Figure~\ref{fig:app-sensitivity}c--d). This TV quantity is an agreement
diagnostic with the HMC baseline; simulation accuracy against known oracle
targets is reported separately.

The deployed rank is selected by the spectral-mass rule without using test
performance or the HMC baseline. This sensitivity study varies rank within
the guided construction; the matched-rank H1 study in
Appendix~\ref{app:sim-guidance-rank} instead compares guided and uniform
direction selection using calibrated-region oracle error.
\fi

The full numerical sensitivity tables are included in the separate
experimental results.

\subsection{Sensitivity to the HMC Baseline Configuration}
\label{app:hmc-strong}
\label{app:tv-caveat}

The main tables use the fixed-budget HMC configuration specified in
Appendix~\ref{app:baselines}. To assess whether its reported predictive
summaries are specific to that configuration, we additionally reran four
representative UCI benchmarks using diagonal VI-based preconditioning and
three jittered chains, retaining the same per-chain trajectory budget.

Across these four benchmarks, the alternative HMC configuration changes
mean test NLL by at most $\numHmcStrongNllShiftMaxAbs$ nats relative to
the primary HMC summaries. Independently calibrated coverage differs by at
most $0.016$, and region size by at most approximately $5\%$.
These comparisons are descriptive because the primary and alternative
summaries use different run cohorts. They show that the predictive quantities
used in our benchmark comparison are similar under this alternative HMC
configuration. The score-guidance claims themselves are evaluated separately
against the simulation oracle targets in Appendices~\ref{app:sim-guidance-rank}
and~\ref{app:bwc-combined}.

\begin{table}[htbp]
\centering
\caption{\textbf{Sensitivity to the HMC baseline configuration.}
The main HMC columns are the 20-seed summaries used in the primary tables.
The alternative configuration uses diagonal VI-based preconditioning and
three jittered chains with the same per-chain trajectory budget; $n$ gives
the number of alternative-configuration runs. Comparisons are descriptive
because the two summaries use different run cohorts. Chain time is per
alternative HMC chain.}
\label{tab:app-hmc-strong}

\small
\setlength{\tabcolsep}{6pt}
\begin{tabular}{l r rr rr rr r}
\toprule
& & \multicolumn{2}{c}{NLL}
& \multicolumn{2}{c}{Coverage}
& \multicolumn{2}{c}{Region size}
& Alt. chain \\
\cmidrule(lr){3-4}\cmidrule(lr){5-6}\cmidrule(lr){7-8}
Benchmark & $n$ & Main & Alt. & Main & Alt. & Main & Alt. & s \\
\midrule
Concrete   & 10 & 2.944 & 2.953 & 0.890 & 0.901 & 15.90 & 15.90 & 1434 \\
Energy     & 10 & 0.881 & 0.883 & 0.895 & 0.879 & 1.63  & 1.57  & 1294 \\
Yacht      & 10 & 1.693 & 1.715 & 0.898 & 0.893 & 4.42  & 4.64  & 1266 \\
Supercond. & 10 & 3.553 & 3.548 & 0.896 & 0.898 & 32.70 & 32.59 & 1559 \\
\bottomrule
\end{tabular}
\end{table}

\subsection{Simulation Oracle Comparisons}
\label{app:sims}

Table~\ref{tab:app-sims} compares predictives with the known conditional
laws. On Simulation~1, SCOPE, HMC, MALA and MFVI have similar mean TV
(approximately $0.07$). On Simulation~2, SCOPE has the smallest mean TV
and KL among the compared methods
($\numTvTrueSimTwoScope$, $\numKlTrueSimTwoScope$), versus
$\numTvTrueSimTwoHmc$, $\numKlTrueSimTwoHmc$ for HMC. Region symmetric
differences are $\numSymDiffSimTwoScope$ and $\numSymDiffSimTwoHmc$.
These comparisons distinguish agreement with HMC from agreement with
the data-generating law.
\begin{table}[htbp]\centering
\caption{\textbf{Oracle comparisons on the simulations (20 seeds).} TV and KL of the test predictive to the true conditional law, RMSE of the predictive mean and standard deviation to their oracle values, and the symmetric difference between each method's calibrated region and the population global-level oracle region (SCOPE: guide-reused region; others: split-conformal on $\Dconf$), with realised coverage.}
\label{tab:app-sims}
\small\setlength{\tabcolsep}{3pt}\renewcommand{\arraystretch}{1.05}
\resizebox{\linewidth}{!}{%
\begin{tabular}{l rrrrrr rrrrrr}\toprule
 & \multicolumn{6}{c}{Simulation 1} & \multicolumn{6}{c}{Simulation 2}\\ \cmidrule(lr){2-7}\cmidrule(lr){8-13}Method & TV & KL & RMSE $\mu$ & RMSE $\sigma$ & sym.\ diff.\ & cov.\ & TV & KL & RMSE $\mu$ & RMSE $\sigma$ & sym.\ diff.\ & cov.\\
\midrule
MFVI & 0.065 & 0.020 & 0.062 & 0.042 & 0.112 & 0.891 & 0.276 & 0.400 & 0.647 & 0.322 & 1.204 & 0.902\\
Laplace & 0.155 & 0.142 & 0.164 & 0.066 & 0.264 & 0.895 & 0.276 & 0.315 & 0.622 & 0.308 & 1.353 & 0.904\\
Trajectory Gaussian (Adam; SWAG-style) & 0.166 & 0.161 & 0.177 & 0.063 & 0.293 & 0.894 & 0.354 & 0.858 & 0.717 & 0.374 & 1.486 & 0.903\\
Trajectory-PCA subspace MCMC & 0.207 & 0.287 & 0.196 & 0.085 & 0.351 & 0.895 & 0.404 & 1.279 & 0.796 & 0.424 & 1.713 & 0.902\\
Deep ensemble & 0.145 & 0.118 & 0.140 & 0.059 & 0.247 & 0.892 & 0.236 & 0.289 & 0.479 & 0.246 & 1.012 & 0.907\\
SGLD & 0.162 & 0.161 & 0.167 & 0.071 & 0.271 & 0.895 & 0.321 & 0.656 & 0.677 & 0.352 & 1.302 & 0.903\\
Full MALA & 0.068 & 0.022 & 0.066 & 0.038 & 0.119 & 0.895 & 0.311 & 0.605 & 0.669 & 0.348 & 1.234 & 0.899\\
Full HMC & 0.066 & 0.022 & 0.062 & 0.039 & 0.117 & 0.894 & 0.213 & 0.231 & 0.470 & 0.226 & 0.811 & 0.906\\
\textbf{SCOPE} & 0.070 & 0.023 & 0.062 & 0.045 & 0.117 & 0.902 & 0.209 & 0.210 & 0.514 & 0.249 & 0.817 & 0.904\\
\bottomrule
\end{tabular}}
\end{table}

\subsection{Run Diagnostics}
\label{app:diagnostics}

Table~\ref{tab:app-diag} reports selected ranks, acceptance, inference
timings, guard scales and energy ratios. The guard diagnostics are
necessary to interpret the difference between the Gaussian completion
identity and the distribution used in the experiments.
\begin{table}[htbp]\centering
\caption{\textbf{SCOPE run diagnostics (means over 20 seeds).} Final active rank $r$ (spectral-mass rule, capped at 80), main-chain acceptance, and wall-clock seconds for one fit. VI, pilot, main and completion show selected components; the fit total also includes active-geometry construction and other fit overhead. End-to-end inference time adds converged-MAP construction and the second candidate fit, charging shared VI once; common MAP initialisation and downstream evaluation are excluded. Timing is reported only for UCI tasks; the simulation candidate fits and frozen baselines were measured under different machine load. Energy ratio and guard scale are means over seeds of their within-fit medians: $D_s(1)/(d_\perp/2)$ before the shrink-only guard and $\gamma_s$, respectively.}
\label{tab:app-diag}
\small\setlength{\tabcolsep}{3pt}\renewcommand{\arraystretch}{1.05}
\begin{tabular}{l rr rrrrr r rr}\toprule
Benchmark & rank $r$ & accept.\ & VI & pilot & main & compl.\ & fit total & end-to-end & energy ratio & guard $\gamma$\\
\midrule
Sim.~1 & 23.4 & 0.59 & -- & -- & -- & -- & -- & -- & 0.87 & 1.00\\
Sim.~2 & 80.0 & 0.58 & -- & -- & -- & -- & -- & -- & 0.63 & 1.00\\
Concrete & 16.4 & 0.59 & 9.4 & 2.5 & 29.3 & 0.5 & 43.1 & 83.3 & 1.86 & 0.96\\
Energy & 7.0 & 0.58 & 8.7 & 2.3 & 28.2 & 0.6 & 40.9 & 77.2 & 25.10 & 0.21\\
Yacht & 3.6 & 0.58 & 8.2 & 2.2 & 26.1 & 0.4 & 37.3 & 71.2 & 30.04 & 0.21\\
Wine & 40.4 & 0.59 & 11.2 & 3.0 & 34.7 & 2.7 & 58.0 & 114.8 & 1.55 & 0.96\\
Abalone & 20.4 & 0.59 & 11.8 & 3.4 & 40.6 & 2.9 & 63.5 & 117.7 & 10.28 & 0.47\\
Supercond. & 18.1 & 0.59 & 10.0 & 2.9 & 31.7 & 1.2 & 48.6 & 94.7 & 0.57 & 1.00\\
Airfoil & 17.6 & 0.59 & 9.0 & 2.4 & 28.8 & 0.8 & 42.9 & 82.7 & 10.13 & 0.40\\
Naval & 10.5 & 0.58 & 9.3 & 2.8 & 32.7 & 1.2 & 48.1 & 90.4 & 18.43 & 0.24\\
Protein & 12.8 & 0.59 & 9.9 & 2.7 & 32.2 & 1.1 & 48.0 & 91.4 & 18.98 & 0.45\\
\bottomrule
\end{tabular}
\end{table}

\subsection{Secondary Outputs and Additional Benchmarks}
\label{app:temperature}
\label{app:excluded}

SCOPE/MHDR selects a scalar $c$ by guide NLL, rescales each component's
deviation from the predictive mean and its likelihood scale by $c$, and
reports a model-mass highest-density region. It is distinct from the
primary untempered guide-reused region. Temperature and tempered-level
comparisons are retained in the result files; they do not provide an
additional validity guarantee for guide reuse.

Airfoil, Naval and Protein use the same evaluation protocol and appear in
the full NLL, coverage and size tables. Their SCOPE coverages are
$\numCovAirfoil$, $\numCovNaval$ and $\numCovProtein$. Naval's size ratio
$\numSizeRatioNaval$ compares regions of order $10^{-3}$, so its relative
difference should be interpreted with the absolute scale.

\subsection{Limitations and Historical Control}
\label{app:discussion}
\label{app:historical}

The slice chain does not target the exact active marginal of the full
posterior. The complement uses diagonal VI scales and an energy guard;
Proposition~\ref{prop:gaussian} characterises only the unguarded construction.
The geometry is local and coverage-level dependent. Finite-dimensional
ESS summaries do not establish full-posterior convergence. The guide-size study provides empirical coverage
comparisons; finite-sample validity requires the independent calibration
used by SCOPE/Split. The Adam-trajectory Gaussian and PCA baselines are
not canonical SGD-based SWAG implementations.

Frozen CABI is retained as a historical control. It uses the VI-mean
anchor without selection, the initial score-kernel weights without pilot
refinement, an unconstrained antithetic VI-scale completion, and a
model-mass level. Its results help separate the full SCOPE construction
from this earlier specification; they do not isolate a single component.


\subsection{Additional mechanism controls for boundary weighting}
\label{app:bwc-combined}

\paragraph{Goal.}
These additional controlled experiments isolate the mechanism behind the
main-simulation result of Appendix~\ref{app:sim-guidance-r4}. We test whether
boundary weighting improves SCOPE's \emph{final calibrated posterior-predictive
region} beyond uniform weighting and random curvature-whitened directions. The
target is region-approximation accuracy, not increased score sensitivity or
chain variability.

\paragraph{Why two experiments.}
Both settings distinguish uncertainty affecting the prediction boundary from
uncertainty mainly affecting the interior density.
\textbf{Experiment A isolates the mechanism:} a rotated 24-parameter density
has two boundary-localized perturbations and 22 interior perturbations. Its
factorized posterior is numerically integrated, avoiding reference-chain error;
40 fresh datasets test direction selection and stronger distributional controls.
\textbf{Experiment B tests a different likelihood:} an eight-parameter Gaussian
mixture combines six interior bumps with a separated shoulder whose inclusion
in the 90\% region changes across three regimes. Its 56 fresh datasets use a
finite full-space MALA reference, not an exact posterior calculation.

\paragraph{Matched comparison.}
At rank two, direction controls share the train-fitted anchor, curvature metric,
main-chain budget, projected VI completion and energy guard. Each predictive
receives its own threshold on the same independent calibration observations. The reference region $C_{\rm ref}$ is built in the same way from the reference
predictive (the integrated posterior predictive in A, the finite full-space
MALA predictive in B), with the same calibration observations.
For the method's region $\widehat C_m$ and reference region $C_{\rm ref}$, error is
\begin{equation}
 d_m=\mathbb E_X\!\left[
 \operatorname{Leb}\{\widehat C_m(X)\mathbin{\triangle}C_{\rm ref}(X)\}
 \right].
 \label{eq:bwc-distance}
\end{equation}
Thus $d_m$ is the expected length of the responses that lie in exactly one of
the two regions; it is zero only when they coincide and, unlike coverage, it
detects a region of the right length in the wrong place. 

For Experiment~A, let $d_{m,j}^{\mathrm{MAP}}$ and
$d_{m,j}^{\mathrm{VI}}$ denote the errors in~\eqref{eq:bwc-distance} for
dataset $j$ at the MAP and VI anchors. Its reported per-dataset error is
\[
E_{m,j}
=
\frac{1}{2}
\left(
d_{m,j}^{\mathrm{MAP}}+d_{m,j}^{\mathrm{VI}}
\right),
\]
in response units.

Experiment~B has no input variable, so for dataset $j$ its absolute
region-disagreement error is
\[
d_{m,j}
=
\operatorname{Leb}\!\left(
\widehat C_{m,j}\mathbin{\triangle}C_{{\rm ref},j}
\right).
\]
Its reported per-dataset error is instead the normalized quantity
\[
E_{m,j}
=
100\,
\frac{d_{m,j}}
{\operatorname{Leb}(C_{{\rm ref},j})},
\]
the percentage of the reference region's length on which the two regions
disagree.

For either experiment, let
\[
\bar E_m=\frac{1}{J}\sum_{j=1}^{J}E_{m,j},
\]
where $J=40$ in Experiment~A and $J=56$ in Experiment~B. Thus the table
reports the mean of the per-dataset errors; in particular, Experiment~B
averages the within-dataset percentages rather than taking a ratio after
averaging across datasets.

For comparator $c$, define the relative error reduction of the guided
construction as
\begin{equation}
\mathrm{Red}_c
=
100\left(
1-\frac{\bar E_{\mathrm{guided}}}{\bar E_c}
\right).
\label{eq:bwc-reduction}
\end{equation}
Hence $\mathrm{Red}_c>0$ means that the guided construction has smaller mean
region-approximation error, whereas $\mathrm{Red}_c<0$ means that it has
larger error. Reductions are computed from the unrounded means.

The experiments are analysed separately; matched main-chain budgets do not
imply equal end-to-end runtime.

\begin{table}[!htbp]
\centering
\caption{\textbf{Boundary weighting reduces final calibrated-region error.}
Mean error $\bar E_m$ is the across-dataset mean of the per-dataset error
defined above: symmetric-difference length in response units for A and
within-dataset symmetric-difference length as a percentage of the reference
region length for B. Paired differences are guided minus comparator, so
negative values favor guided inference. A uses Bonferroni-adjusted paired
bootstrap intervals targeting 95\% familywise coverage for its four primary
contrasts ($\dagger$: two secondary contrasts); B uses pointwise 95\% paired
bootstrap intervals. The final column reports
$\mathrm{Red}_c=100(1-\bar E_{\mathrm{guided}}/\bar E_c)$, the percentage
reduction in mean region-approximation error of the guided construction
relative to the row's comparator; positive values favor guided inference.
All reductions use unrounded means. A printed $-0.00$ does not establish
exclusion of zero.}

\label{tab:bwc-combined-results}
\small
\setlength{\tabcolsep}{3pt}
\begin{tabular}{@{}l r l r@{}}
\toprule
Construction & Mean error $\bar E_m$ & Paired difference interval &
Relative error reduction (\%)\\
\midrule
\multicolumn{4}{@{}l}{\textbf{A. Integrated posterior reference; 40 datasets; absolute error}}\\
\textbf{Boundary-weighted SCOPE} & \textbf{0.00579} & -- & --\\
Uniform weights & 0.01779 & $[-0.01517,-0.00869]$ & $67.4\%$\\
Whitened random span & 0.01990 & $[-0.01736,-0.01074]$ & $70.9\%$\\
Permuted boundary weights & 0.01819 & $[-0.01552,-0.00912]$ & $68.1\%$\\
Same-span oracle Gaussian$^\dagger$ & 0.01316 & $[-0.00938,-0.00525]$ & $56.0\%$\\
\midrule
\multicolumn{4}{@{}l}{\textbf{B. Finite-MALA reference; 56 datasets; normalized error (\%)}}\\
Boundary kernel + refinement & 0.28 & -- & --\\
Kernel only & \textbf{0.23} & $[-0.00,+0.13]$ & $-21\%$\\
Uniform weights & 0.70 & $[-0.74,-0.17]$ & $60\%$\\
Whitened random span & 0.82 & $[-0.87,-0.29]$ & $65\%$\\
\bottomrule
\end{tabular}
\end{table}

\paragraph{What the experiments show.}
\textbf{Boundary weighting contributes beyond whitening and arbitrary span
selection.} In A, mean error falls by 67.4\% against uniform weights and
70.9\% against random spans. Permuting the weights within gradient-norm
strata preserves their values but increases error: guided inference is 68.1\%
better, isolating the importance of \emph{weight placement}. Its 56.0\%
improvement over a same-span Gaussian with exact slice mean and covariance
shows that moment matching alone does not recover the relevant predictive
integral. Moving the same perturbations into the interior preserves the
posterior but removes the region-fidelity advantage.

\textbf{The mixture experiment corroborates the benefit:} guided error is
60\% lower than uniform and 65\% lower than random. Completion improves guided
error but worsens uniform and random error, linking the gain to the division
of uncertainty between active sampling and the approximate complement.

\paragraph{Conclusion.}
\textbf{Boundary-sensitive direction selection measurably improves the final
calibrated posterior-region approximation at low active rank.}
A's largest uniform-weight gains occur at MAP, but method-specific guide
selection retains reductions of 53.8\% and 65.5\% against uniform and random.
The evidence supports boundary localisation, not an additional benefit of
pilot refinement, accuracy dominance under automatic rank selection, or
better true-law prediction. It directly supports SCOPE's allocation principle:
\emph{spend non-Gaussian inference on uncertainty that matters to the region}.
These mechanism studies complement the main-simulation rank sweep in
Appendix~\ref{app:sim-guidance-rank}, where the downstream advantage is largest
at low rank and decreases as more active directions are included.

\end{document}